\documentclass[9pt,shortpaper,twoside,web]{ieeecolor}
\usepackage{generic}
\usepackage{cite}
\usepackage{amsmath,amssymb,amsfonts}
\usepackage{algorithmic}
\usepackage{graphicx}
\usepackage{textcomp}
\usepackage{multirow}
\usepackage{booktabs}

\usepackage{epstopdf}
\usepackage{bm}
\usepackage{multirow}
\usepackage{etoolbox}
\makeatletter
\patchcmd{\@makecaption}
  {\scshape}
  {}
  {}
  {}
\makeatletter
\patchcmd{\@makecaption}
  {\\}
  {:\ }
  {}
  {}
\makeatother

\def\BibTeX{{\rm B\kern-.05em{\sc i\kern-.025em b}\kern-.08em
    T\kern-.1667em\lower.7ex\hbox{E}\kern-.125emX}}
\begin{document}
\title{\huge Brain Imaging-to-Graph Generation using Adversarial Hierarchical Diffusion Models for MCI Causality Analysis}
\author{Qiankun Zuo, Hao Tian, Chi-Man Pun, Hongfei Wang, Yudong Zhang, Jin Hong
\thanks{Corresponding author: Yudong Zhang, yudongzhang@ieee.org; Jin Hong, hongjin@ncu.edu.cn}
\thanks{Qiankun Zuo and Hao Tian are with the School of Information Engineering, Hubei University of Economics, Wuhan 430205, Hubei, China.}
\thanks{Chi-Man Pun is with the Department of Computer and Information Science, University of Macau, Macao 519000, China.}
\thanks{Hongfei Wang is with the Department of Orthopaedics and Traumatology, University of Hong Kong, Hong Kong 999077, China.}
\thanks{Yudong Zhang is with the School of Computing and Mathematic Sciences, University of Leicester, Leicester LE1 7RH, United Kingdom.}
\thanks{Jin Hong is with the School of Information Engineering, Nanchang University, Nanchang, 330031, China.}
}

\maketitle

\begin{abstract}
Effective connectivity can describe the causal patterns among brain regions. These patterns have the potential to reveal the pathological mechanism and promote early diagnosis and effective drug development for cognitive disease. However, the current methods utilize software toolkits to extract empirical features from brain imaging to estimate effective connectivity. These methods heavily rely on manual parameter settings and may result in large errors during effective connectivity estimation.
In this paper, a novel brain imaging-to-graph generation (BIGG) framework is proposed to map functional magnetic resonance imaging (fMRI) into effective connectivity for mild cognitive impairment (MCI) analysis.
To be specific, the proposed BIGG framework is based on the diffusion denoising probabilistic models (DDPM), where each denoising step is modeled as a generative adversarial network (GAN) to progressively translate the noise and conditional fMRI to effective connectivity.
The hierarchical transformers in the generator are designed to estimate the noise at multiple scales. Each scale concentrates on both spatial and temporal information between brain regions, enabling good quality in noise removal and better inference of causal relations.
Meanwhile, the transformer-based discriminator constrains the generator to further capture global and local patterns for improving high-quality and diversity generation.
By introducing the diffusive factor, the denoising inference with a large sampling step size is more efficient and can maintain high-quality results for effective connectivity generation.
Evaluations of the ADNI dataset demonstrate the feasibility and efficacy of the proposed model. The proposed model not only achieves superior prediction performance compared with other competing methods but also predicts MCI-related causal connections that are consistent with clinical studies.
\end{abstract}


\section{Introduction}
\label{sec:introduction}

\IEEEPARstart{B}rain effective connectivity (BEC) describes the causal interaction from one brain region to another and helps perform various cognitive and perceptual tasks \cite{reid2019advancing}. It aims to transmit and analyze functional information between distant regions of the brain with the characteristics of high speed and precision \cite{avena2018communication}. Recent studies have shown that the abnormal changes of BEC can reflect the underlying pathology of brain diseases \cite{rupprechter2020blunted,mijalkov2022directed}. These changes are probably accompanied by alterations in the brain's microstructure \cite{hampstead2016patterns,sami2018neurophysiological}, such as those associated with Alzheimer's disease (AD) and its early-stage mild cognitive impairment (MCI). Investigation of the BEC is helpful for researchers to understand the underlying mechanisms associated with neurodegenerative diseases and develop potential treatments or new drugs for rehabilitation \cite{shi2019leveraging,scherr2021effective}. Therefore, constructing BEC from brain imaging (i.e., functional magnetic resonance imaging, fMRI) becomes a very promising way to analyze cognitive disease and identify possible biomarkers for MCI diagnosis.

Constructing BEC involves learning a mapping network to predict directional connections from one neural unit to another by analyzing their functional signals. Since the fMRI has the advantages of being noninvasive and having high temporal resolution, it has drawn great attention to extracting the complex connectivity features for brain disease diagnosis\cite{pei2022data}. The brain functional connectivity (BFC) gives the temporal correlation between any pair of neural units, while the BEC is an asymmetric matrix representing the directional information of neural transmission. The directed graph can be used to analyze BEC, including a set of vertices (named regions-of-interest, or ROIs) and a set of directed edges (effective connections). Previous researchers focused on BEC estimation by using traditional learning methods, including the dynamic causal models (DCM) \cite{friston2014dcm}, the Bayesian network (BN) method \cite{liu2019learning}, the correlation analysis method \cite{xu2017initial}, and so on. These methods utilize shallow network structures or prior knowledge, which are unable to extract complex causal features from fMRI and bring low performance in disease analysis.

Deep learning methods have been widely applied in the exploration of BEC estimation \cite{al2021severity,bagherzadeh2022detection}. They not only achieve excellent performance on image recognition tasks in Euclidean space but also show good results on brain network generation in non-Euclidean space. The primary characteristic is the strong ability to perform high-level and complex feature extraction. Increasingly new methods based on deep learning have been explored to construct BEC from functional MRI data \cite{liu2021inferring,zhang2022estimating,ji2023dynamic}. However, the current methods heavily rely on the software toolkit to preprocess the fMRI for extracting empirical features (ROI-based time-series). The main drawback is that the manual parameter settings from different researchers may result in large errors when using these empirical features to estimate BEC.

As the most popular and powerful generative model, the generative adversarial network (GAN) \cite{goodfellow2020generative} implicitly characterizes the distribution of synthetic samples through a two-player adversarial game. It can generate high-quality samples with efficient computation while producing homogeneous samples because of training instability and mode collapse. An alternative way to solve this issue is the emergence of diffusion denoising probabilistic models (DDPM) \cite{ho2020denoising} that have received great attention in generating tasks\cite{wang2024srfs,zhang2024anomaly}. The main advantage of DDPM is its powerful ability to generate high-quality and diverse samples, and the disadvantage lies in its expensive computation.
Inspired by the above observations, we intend to combine the advantages of the GAN and the DDPM for generation performance improvement. A novel brain imaging-to-graph generation (BIGG) model is proposed to generate BEC from fMRI to analyze MCI causality. Specifically, the DDPM is split into several diffusive steps, and each step is modeled as a conditional GAN to denoise the Gaussian sample and generate clean samples. The transformer-based generator and discriminator are combined to capture hierarchical spatial-temporal features for noise removal. During denoising inference, a diffusive factor is introduced to increase sampling step size for efficient and high-quality sample generation. As a result, the estimated BEC reflects a more sophisticated causal relationship between brain regions and captures the MCI-related features. To the best of our knowledge, the proposed BIGG is a unified AIGC framework that firstly translates fMRI into BEC. The main contributions of this work are summarized as follows:

\begin{itemize}
	
	\item The proposed BIGG model is a unified framework to map fMRI onto effective connectivity in an end-to-end manner. It leverages the adversarial strategy to model each denoising step of DDPM by introducing conditional fMRI, which is high-quality, diverse, and efficient for effective connectivity generation.
	
	\item The hierarchical denoising transformer is designed to learn multi-scale features for noise removal. By focusing on both spatial and temporal domains, the quality of denoising process is greatly improved, and the performance of causal inference is significantly enhanced.

	\item The multi-resolution consistent transformer is devised to approximate the temporal distribution of diffusive samples to that of denoised samples at different scales. It promotes the generator to capture global and local patterns for good generation in terms of diversity and stability.

\end{itemize}

The rest of this paper is structured as follows: The related works are introduced in Section~\ref{s2}. The overall architecture of the proposed BIGG model is presented in Section~\ref{s3}. The experimental results, including generation evaluation and classification performance, are described in Section~\ref{s4}. The reliability of our results is discussed in Section~\ref{s5}, and Sections~\ref{s6} draw the main conclusions.

\begin{figure}[t]
	\centering
	\includegraphics[width=\columnwidth]{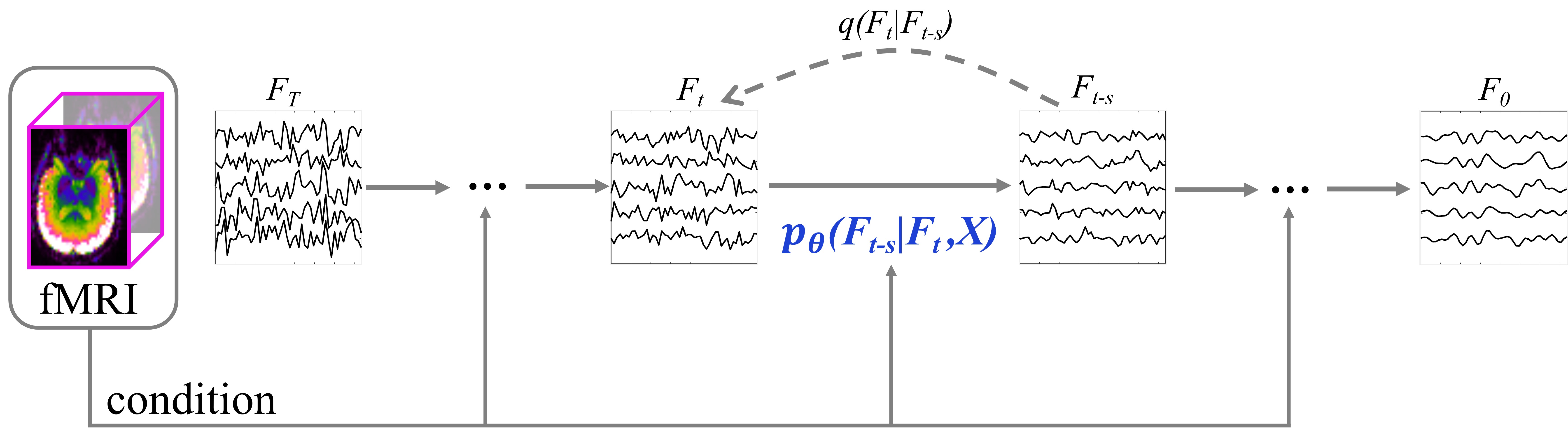}
	\caption{The architecture of the proposed BIGG framework. In the forward process, the empirical sample $F_0$ is transformed to the normal Gaussian noisy sample $F_T$ by gradually adding noise. In the reverse process, the fMRI is considered as a condition to guide the network $p_ \theta$ to remove the noise from $F_T$. \label{fig0}}
\end{figure}

\section{Related Work}
\label{s2}
\subsection{BEC learning methods}
Functional connectivity network analysis is usually used to identify abnormal connectivity patterns that are associated with different cognitive functions, such as memory, attention, and emotion. The BEC belongs to the functional connectivity network and can bridge causal connections between brain regions. Many studies have focused on exploring the BEC for better diagnostic performance and good interpretability. There are two categories of methods for learning BEC from functional data: shallow learning methods and deep learning methods.

The frequently used shallow method is the dynamic causal model (DCM)\cite{friston2014dcm}. For example, Park et al.\cite{park2018dynamic} employed the parametric empirical Bayes method to model the directed effects of sliding windows. And the Granger causality (GC)\cite{seth2010matlab,seth2015granger} is the most commonly used shallow method. For example, DSouza et al. \cite{dsouza2017exploring} utilize the multiple regression algorithm to process historical information from functional time series for causal interaction prediction. These methods cannot extract deep and complex connectivity features from fMRI.

To explore the deep features of brain imaging, deep learning methods show great success in causal modeling between brain regions. The work in \cite{talebi2019ncreann} employed nonlinear causal relationship estimation with an artificial neural network to predict causal relations between brain regions by analyzing both linear and nonlinear components of Electroencephalogram (EEG) data. Also, Abbasvandi et al.\cite{abbasvandi2019self} combined the recurrent neural network and Granger causality to estimate effective connectivity from EEG data. They greatly improved the prediction accuracy of the simulation data and the epileptic seizure dataset. Considering the great ability of GANs to characterize data distribution, Liu et al.\cite{liu2020ec} designed a GAN-based network to infer directed connections from fMRI data. To capture temporal features, they\cite{ji2021estimating} employed recurrent generative adversarial networks for effective connectivity learning. Presently, Zou et al.\cite{zou2022exploring} introduced the graph convolutional network (GCN) to mine both temporal and spatial topological relationships among distant brain regions for learning BECs. Although the above deep learning methods achieved promising prediction performance in BEC estimation, they heavily rely on the software toolkit to preprocess the fMRI for extracting empirical time-series data. That may result in large errors due to different manual parameter settings during preprocessing procedures.

\subsection{Generative learning models}

Generative adversarial networks (GANs) have ruled generative approaches since they were first proposed by Goodfellow. The primary advantage is implicitly modeling the distribution of the generated samples through a two-player game. Many variants of GANs have been proposed and applied to many generation tasks, such as generating super-resolution, synthesizing cross-modal data, segmenting images, and so on. To satisfy specific generating tasks, conditional GAN\cite{mirza2014conditional} and related variants also achieve quite good performance efficiently\cite{wang20183d}. The problem of instability and mode collapse in training has not been completely addressed yet. This may lead to homogeneously generated data and hinder its wider applications. Recently, denoising diffusion probabilistic models (DDPM) have attracted much attention in image generation \cite{ozdenizci2023restoring} because of their ability to generate high-quality and diverse samples. DDPM aims to denoise the Gaussian sample gradually and recover the clean sample. However, the denoising process requires a Gaussian distribution assumption with only a small denoising step, which leads to slow reverse diffusion in about thousands of steps to approach clean data. Based on the above observations, we try to combine the advantages of GAN and DDPM in generation, such as efficiency, high quality, and diversity. We propose the novel BIGG model to precisely estimate BEC from fMRI for MCI causality analysis.

\begin{figure*}[t]
	\centering
	\includegraphics[width=0.8\textwidth]{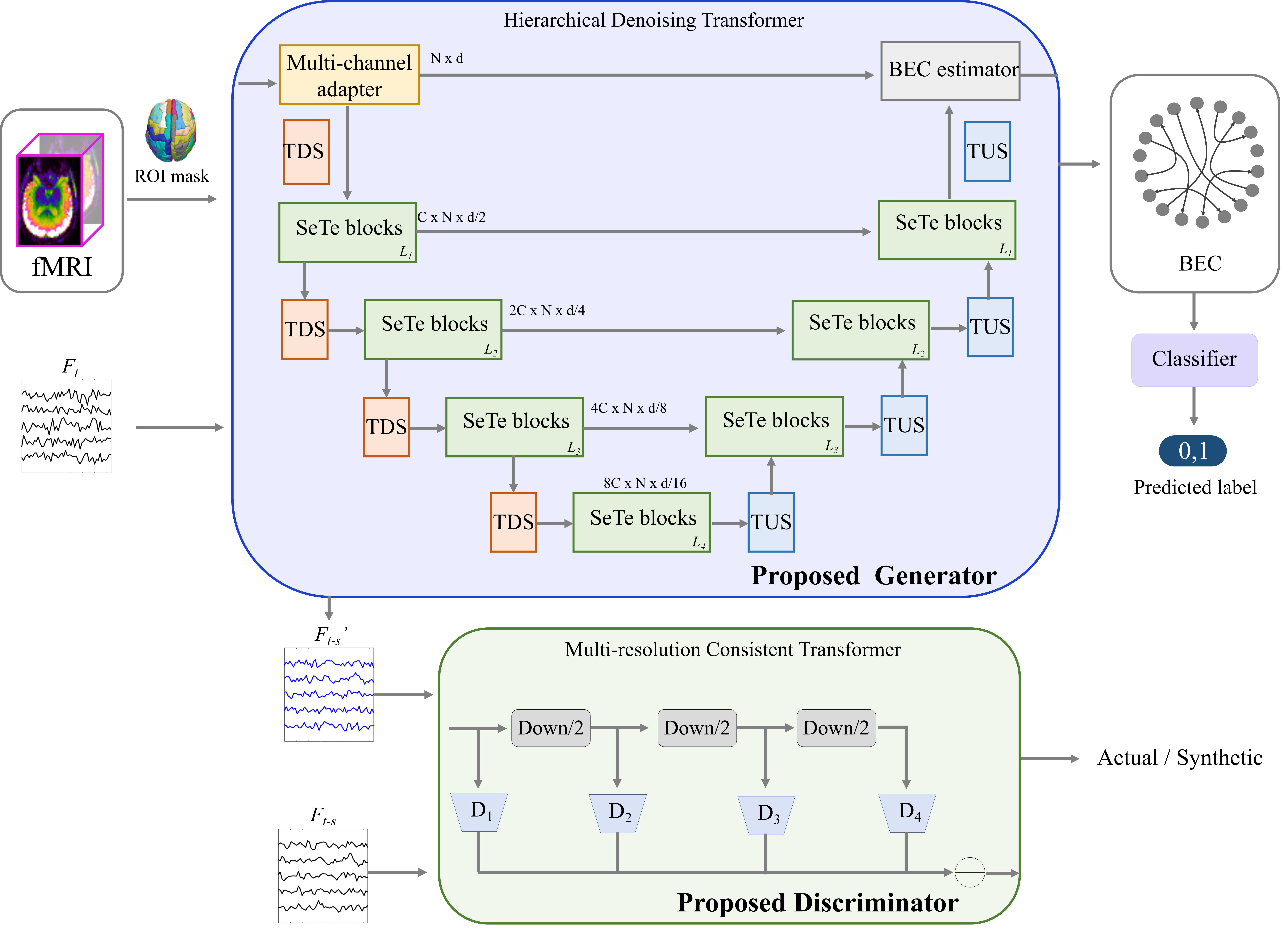}
	\caption{Detailed network structure of one denoising step. It includes one generator, one discriminator, and one classifier. The input is the conditional fMRI and noisy sample $F_t$, and the output is the denosed sample $F_{t-s}'$ and BEC.\label{fig1}}
\end{figure*}

\section{Method}
\label{s3}

As shown in Fig.~\ref{fig0}, the proposed BIGG model is a diffusion-based model to denoise a Gaussian sample into a clean sample. At each denoising step, the fMRI is treated as a condition to guide a GAN-based network to remove noise and estimate BEC. The details of one denoising step are shown in Fig.~\ref{fig1}, including one generator and one discriminator. At the final denoising step, the obtained BEC is used to analyze MCI causality.
In the following, we first introduce conventional diffusion models and then describe the adversarial hierarchical diffusion models with transformer-based generators and discriminators. Finally, hybrid objective functions are devised to optimize the proposed BIGG model.

\subsection{Conventional Denoising Diffusion Model}
The basic principle of the diffusion model is to learn the information attenuation caused by noise and then use the learned noise to generate a clean sample. It is usually divided into the forward process and the inverse process. In the diffusion process, Gaussian noise is constantly added to the input sample $F_0$ for sufficiently large $T$ (hundreds or thousands) steps. Under the rules of the Markov chain, the probability distribution of a noisy sample $F_T$ will approach the stationary distribution (such as the Gaussian distribution) at the $T$-th step. Here is the diffusion formula:

\begin{equation}
q\left(\mathbf{F}_{1: T} \mid \mathbf{F}_{0}\right)=\prod_{t=1}^{T} q\left(\mathbf{F}_{t} \mid \mathbf{F}_{t-1}\right)
\end{equation}
\begin{equation}\label{eq2}
q\left(\mathbf{F}_{t} \mid \mathbf{F}_{t-1}\right)=\mathcal{N}\left(\mathbf{F}_{t} ; \sqrt{1-\beta_{t}} \mathbf{F}_{t-1}, \beta_{t} \mathbf{I}\right)
\end{equation}
where $t = \{1,2,...,T\}$. $\beta_{t}$ is the noise variance that is defined before the model's training. $\mathcal{N}$ is the assumed Gaussian distribution, and $\mathbf{I}$ is an identity matrix. The reverse process also follows the Markov chain to translate the noisy sample $F_T$ to a cleaned sample $F_0$. The assumption of a large step $T$ and a small $\beta_{t}$ can model the denoising probability in a Gaussian distribution:

\begin{equation}\label{eq3}
p_{\theta}\left(\mathbf{F}_{0: T}\right)=p\left(\mathbf{F}_{T}\right) \prod_{t=1}^{T} p_{\theta}\left(\mathbf{F}_{t-1} \mid \mathbf{F}_{t}\right)
\end{equation}
\begin{equation}\label{eq4}
p_{\theta}\left(\mathbf{F}_{t-1} \mid \mathbf{F}_{t}\right)=\mathcal{N}\left(\mathbf{F}_{t-1} ; \boldsymbol{\mu}_{\theta}\left(\mathbf{F}_{t}, t\right), \sigma_{t}^{2} \mathbf{I}\right)
\end{equation}
where $\boldsymbol{\mu}_{\theta}(\mathbf{F}_{t}, t)$ and $\sigma_{t}^{2}$ are the mean and variance of the denoised sample $\mathbf{F}_{t-1}$. $\theta$ indicates the network's parameters.

We use the deep learning model $p_ \theta (\mathbf{F}_{t-1}| \mathbf{F}_{t})$ to approximate the true distribution $q(\mathbf{F}_{t-1}| \mathbf{F}_{t})$, and get the reparameterization of $\mu$ and $\sigma_{t}^{2}$ in the following form:
\begin{equation}
\boldsymbol{\mu}_{\theta}\left(\mathbf{F}_{t}, t\right)=\frac{1}{\sqrt{\alpha_{t}}}\left(\mathbf{F}_{t}-\frac{\beta_{t}}{\sqrt{1-\bar{\alpha}_{t}}} \boldsymbol{\epsilon}_{\theta}\left(\mathbf{F}_{t}, t\right)\right)
\end{equation}
\begin{equation}
\sigma_{t}^{2} = \frac{\left(1-\bar{\alpha}_{t-1}\right)}{\left(1-\bar{\alpha}_{t}\right)} \beta_{t}
\end{equation}
here, $\alpha_{t}=1-\beta_{t}, \bar{\alpha}_{t}=\prod_{i=1}^{t} \alpha_{i}$. By applying a variational evidence lower bound (ELBO) constraint, the added noise $\boldsymbol{\epsilon}_{\theta}(\mathbf{F}_{t}, t)$ can be calculated by minimizing the MSE loss:

\begin{equation}
\mathbb{E}_{t, \mathbf{F}_{0}, \boldsymbol{\epsilon}}\left[\left\|\boldsymbol{\epsilon}-\boldsymbol{\epsilon}_{\theta}\left(\sqrt{\bar{\alpha}_{t}} \mathbf{F}_{0}+\sqrt{1-\bar{\alpha}_{t}} \boldsymbol{\epsilon}, t\right)\right\|^{2}\right]
\end{equation}
where, $\boldsymbol{\epsilon} \sim \mathcal{N}(\mathbf{0},\mathbf{I})$. After the model's optimization, new samples can be derived from Gaussian noise $F_T \sim \mathcal{N}(\mathbf{0},\mathbf{I})$ by reversely diffusing $T$ steps based on Eq.(~\ref{eq3}) and Eq.(~\ref{eq4}).

\subsection{Adversarial Hierarchical Diffusion Models}
Conventional DDPM can generate high-quality samples but suffers from low efficiency in sampling because of the thousands of denoising steps. While the GAN can make up for this shortcoming by having fast-generating ability, Thus, we utilize adversarial hierarchical diffusion models by combining both DDPM and GAN for efficient and high-fidelity sampling generation. Apart from this, there are two other differences compared with the conventional DDPM: (1) fMRI is treated as a condition to guide the denoising process to generate a clean subject-specific sample using our model; the conventional diffusion process likely generates uncontrollable samples that cannot reflect subject-specific disease information; (2) the step size is reduced by a diffusive factor of $s$ ($s>>1$), which speeds the generation process and keeps the sample generation high-quality. Before the diffusion process, the conditional fMRI is parcellated to obtain the rough sample $\mathbf{X}$ using the non-parameter method.
In this method, the voxels in the corresponding brain regions of the fMRI and the anatomical atlas ($N$ ROIs) are dotted and summed, and the output is the ROI-based two-dimensional time-series (rough sample $\mathbf{X}$).

The adversarial hierarchical diffusion models have several denoising steps. Each step adopts a conditional GAN to remove the noise from previous noisy sample. As shown in Fig.~\ref{fig1}, it consists of two parts: one generator and one discriminator.
In the forward direction, the empirical sample $\mathbf{F}_0$ is transformed to the normal Gaussian noisy sample $\mathbf{F}_T$ by gradually adding noise. The computation formula is the same as Eq.(~\ref{eq2}). In the reverse direction, we incorporate the conditional $\mathbf{X}$ into the denoising procedure. First, the rough sample $\mathbf{X}$ is considered as a condition to guide the generator $\mathbf{G}_{\theta}$ predict the initial sample $\mathbf{\hat F}_{0}$ from $\mathbf{F}_T$, then the posterior sampling is utilized to synthesize the denoised sample $\mathbf{F}_{t-s}'$; meanwhile, the transformer-based discriminator $\mathbf{D}_ \theta$ distinguishes the actual ($\mathbf{F}_{t-s}$) or synthetic ($\mathbf{F}_{t-s}'$) for the denoising process. Specifically, at the $t$ step, we aim to predicte $\mathbf{F}_{t-s}$ from $\mathbf{F}_{t}$. Firstly, a generator $\mathbf{G}_{\theta}(\mathbf{F}_t,\mathbf{X},t)$ is utilized to predict the initial sample $\mathbf{\hat F}_0^{[t/s]}$, then $\mathbf{F}_{t-s}$ is sampled using the posterior distribution $q(\mathbf{F}_{t-s}|\mathbf{F}_{t},\mathbf{\hat F}_0^{[t/s]})$ by giving $\mathbf{F}_{t}$ and $\mathbf{\hat F}_0^{[t/s]}$. Finally, after $T/s$ steps, the ultimate denoised sample $\mathbf{F}_0'$ (equal to $\mathbf{\hat F}_0^{[1]}$) sampled from the estimated distribution $p_{\theta}(\mathbf{F}_{0} \mid \mathbf{F}_{s},\mathbf{X})$. The denoising process can be expressed as follows:

\begin{figure}[htbp]
	\centering
	\includegraphics[width=\columnwidth]{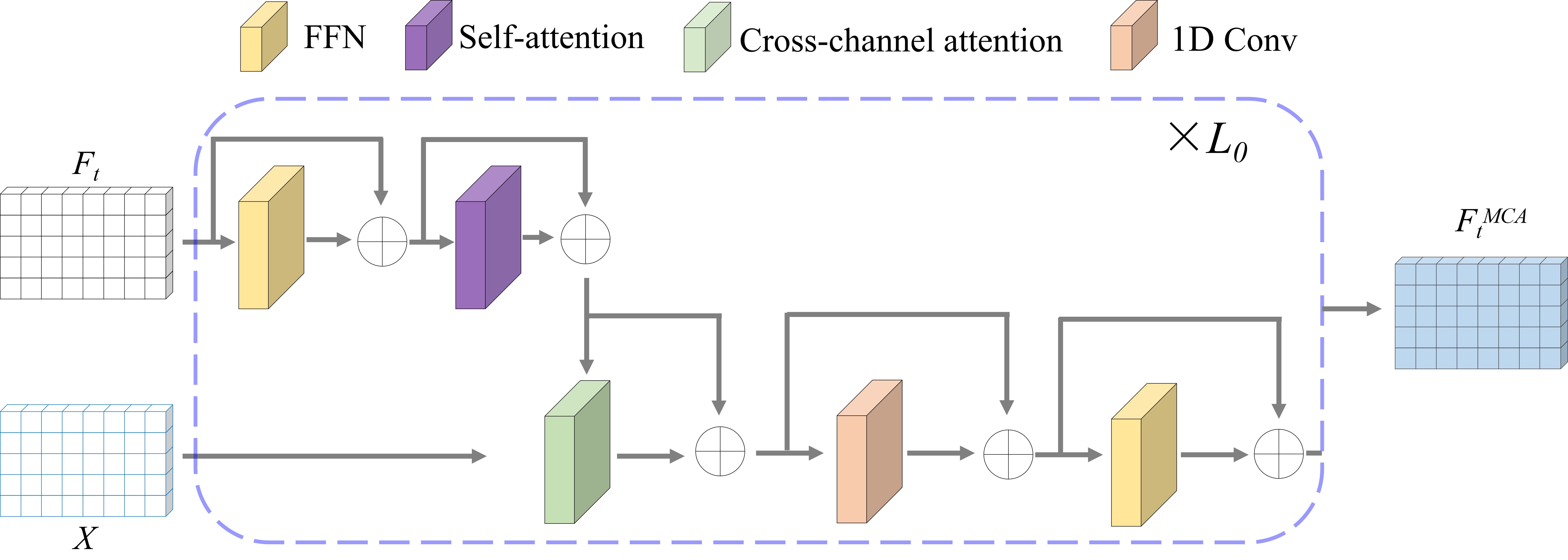}
	\caption{The structure of the multi-channel adaptor. The inputs are the noisy sample $\mathbf{F}_t$ and the rough sample $\mathbf{X}$, and the output is the fused sample. The three samples share the same dimension.\label{fig2}}
\end{figure}

\begin{figure}[h]
	\centering
	\includegraphics[width=\columnwidth]{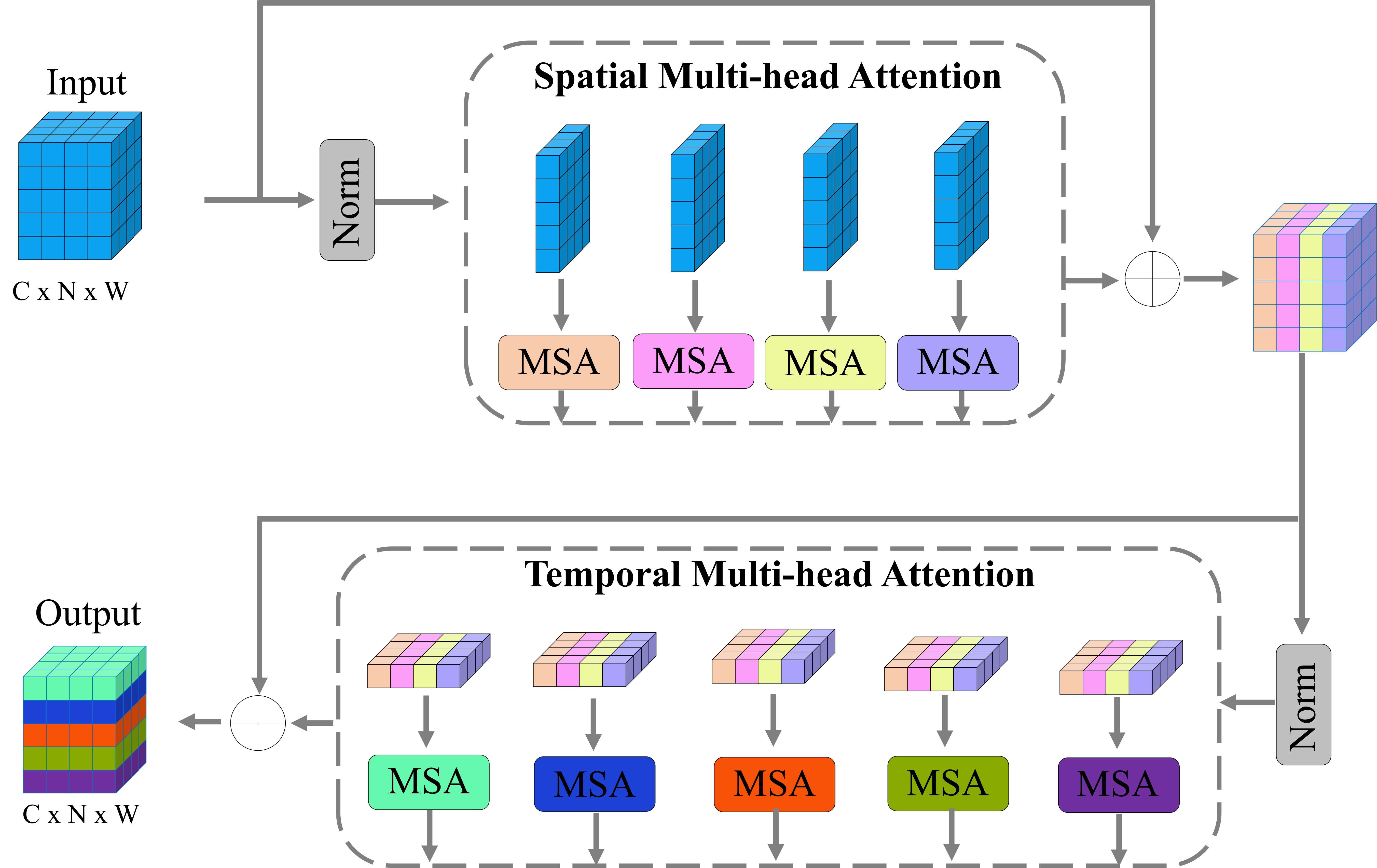}
	\caption{The detailed structure of the SeTe block. The input and output share the same dimension. It passes successively through the spatial multi-head attention module and the temporal multi-head attention module.\label{fig3}}
\end{figure}

\begin{equation}
p_{\theta}\left(\mathbf{F}_{t-s} \mid \mathbf{F}_{t}, \mathbf{X}\right)=q\left(\mathbf{F}_{t-s} \mid \mathbf{F}_{t}, \hat{\mathbf{F}}_{0}^{[t / s]}=\mathbf{G}_{\theta}^{[t / s]}\left(\mathbf{F}_{t}, \mathbf{X}, t\right)\right)
\end{equation}
\begin{equation}
q\left(\mathbf{F}_{t-s} \mid \mathbf{F}_{t}, \hat{\mathbf{F}}_{0}^{[t / s]}\right)=q\left(\mathbf{F}_{t} \mid \mathbf{F}_{t-s}, \hat{\mathbf{F}}_{0}^{[t / s]}\right) \frac{q\left(\mathbf{F}_{t-s} \mid \hat{\mathbf{F}}_{0}^{[t / s]}\right)}{q\left(\mathbf{F}_{t} \mid \hat{\mathbf{F}}_{0}^{[t / s]}\right)}
\end{equation}
The sampling probability from $p_{\theta}\left(\mathbf{F}_{t-s} \mid \mathbf{F}_{t}, \mathbf{X}\right)$ is defined as:
\begin{equation}
\mathbf{F}_{t-s} = \frac{\sqrt{\alpha_{t}}\left(1-\bar{\alpha}_{t-s}\right)}{1-\bar{\alpha}_{t}} \mathbf{F}_{t}+\frac{\sqrt{\bar{\alpha}_{t-s}} \beta_{t}}{1-\bar{\alpha}_{t}} \mathbf{G}_{\theta}^{[t / s]} + \boldsymbol{\epsilon} \sqrt{\beta_{t}}
\end{equation}

\subsubsection{Generator with the hierarchical denoising transformer}

The aim of the transformer-based generator ($\mathbf{G}_{\theta}$) is to remove the noise from noisy samples to obtain clean samples by conditional guidance. Specifically, at $t$-the step, the input is the noisy sample $\mathbf{F}_t$ and the rough sample $\mathbf{X}$, and the output is the initial denoised sample $\mathbf{\hat F}_0$. All of them share the same size, $N \times d$. The generator consists of four modules, including the multi-channel adaptor (MCA), the spatial-enhanced temporal-enhanced (SeTe) blocks, the temporal down- and up-sampling (TDS and TUS), and the brain effective connectivity (BEC) estimator.

The MCA adaptively fuses the noisy sample $\mathbf{F}_t$ and the rough sample $\mathbf{X}$, Different from the traditional way of concatenating the two samples, we designed a cross-channel attention mechanism when fusing them. As shown in Fig.~\ref{fig2}, first, the sample $\mathbf{F}_t$ is passed through the feed-forward network (FFN) and self-attention network; then, it is projected on the $\mathbf{X}$ to compute weighting scores for itself. We denote the input of cross-channel attention as $\mathbf{E}_t$ and $\mathbf{X}$, the output of this fusion operation can be expressed as follows:
\begin{equation}
\mathbf{E}_{t}'=\operatorname{softmax}\left(\frac{\mathbf{E}_t \mathbf{X}^{T}}{\sqrt{d}}\right) \mathbf{E}_t
\end{equation}
Next, the $1 \times 3$ convolutional kernel is applied for each ROI to extract local features. Finally, FFN is used to adjust temporal features among different channels. The output of every sub-network has the same size, $N \times d$. The output of the MCA module is $\mathbf{F}_t^{MCA}$.

The TDS is used to halve the feature dimension and add the channels. Taking the first TDS as an example, for each ROI feature, we apply a 1D convolutional kernel with the size $1 \times 3$ to extract local temporal features, and through $C$ convolutional kernels with a step size of $2$, each ROI feature is translated to a sequence of vectors with the length $d/2$. The final output $\mathbf{F}_{t}^{TDS}$ has the size $C \times N \times d/2$.

\begin{figure}[h]
	\centering
	\includegraphics[width=\columnwidth]{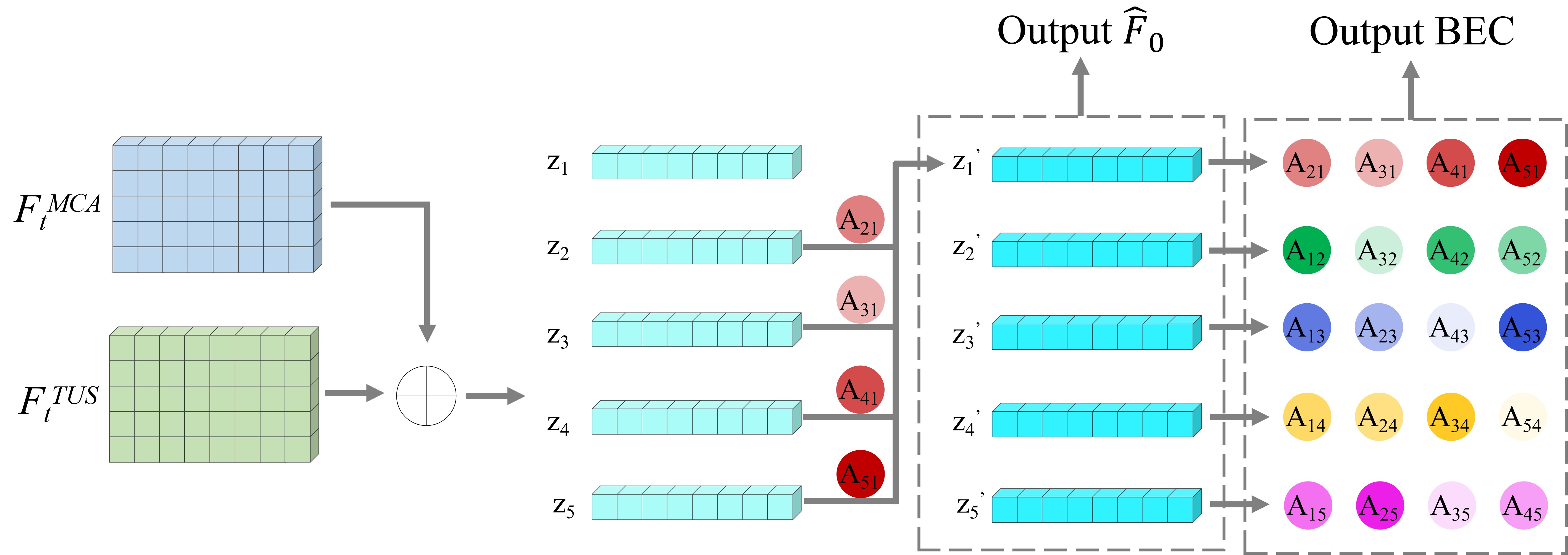}
	\caption{The structure of the BEC estimator. The outputs are a denoised sample and an asymmetric brain connectivity matrix. \label{fig4}}
\end{figure}

The SeTe blocks are designed to extract both spatial and temporal features. The conventional transformed-based method focuses on the spatial correlation between pairs of ROIs while ignoring the temporal continuity. The SeTe benefits from multi-head attention (MSA), which enhances long-term dependence both spatially and temporally. As shown in Fig.~\ref{fig3}, it is comprised of a spatial multi-head attention (SMA) and a temporal multi-head attention (TMA). The difference between these two attention networks is that the former is operated in the spatial direction, and the latter is operated in the temporal direction. The input is a tensor with the size $C \times N \times d/2$, and the output $\mathbf{F}_{t}^{SeTe}$ can be defined by:
\begin{equation}
\mathbf{F}_{t}'^{SeTe}=\mathrm{SMA}\left(\mathrm{Norm}\left(\mathbf{F}_{t}^{TDS}\right)\right)+\mathbf{F}_{t}^{TDS}
\end{equation}
\begin{equation}
	\mathbf{F}_{t}^{SeTe}=\mathrm{TMA}\left(\mathrm{Norm}\left(\mathbf{F}_{t}'^{SeTe}\right)\right)+\mathbf{F}_{t}'^{SeTe}
\end{equation}
here, the $\mathbf{F}_{t}'^{SeTe}$ and $\mathbf{F}_{t}^{SeTe}$ have the same size, $C \times N \times d/2$. Specifically, the SMA splits the input $\mathbf{F}_{t}^{SeTe}$ into several parallel parts and applies MSA to each of them. We denote $H$ as the head number. The detailed computation of SMA is expressed below:
\begin{equation}
	\mathbf{F}_{t}^{SeTe}=\{ \mathbf{F}_{t}^{SeTe(1)},\mathbf{F}_{t}^{SeTe(2)}, \cdots , \mathbf{F}_{t}^{SeTe(d)} \}
\end{equation}
\begin{equation}
	\begin{aligned}
	\mathbf{\hat F}_{t}^{SeTe(i)} &= \operatorname{Att}\left(\mathbf{F}_{t}^{SeTe(i)} \mathbf{W}_{h}^{Q}, \mathbf{F}_{t}^{SeTe(i)} \mathbf{W}_{h}^{K}, \mathbf{F}_{t}^{SeTe(i)} \mathbf{W}_{h}^{V}\right) \\
	&=\operatorname{Att}(\mathbf{Q}_h^i,\mathbf{K}_h^i,\mathbf{V}_h^i) \\
	&=\operatorname{softmax}\left(\frac{\mathbf{Q}_h^i (\mathbf{K}_h^i)^{T}}{\sqrt{C/H}}\right) \mathbf{V}_h^i
	\end{aligned}
\end{equation}
\begin{equation}
\mathbf{F}_{t}'^{SeTe} = \{ \mathbf{\hat F}_{t}^{SeTe(1)},\mathbf{\hat F}_{t}^{SeTe(2)}, \cdots , \mathbf{\hat F}_{t}^{SeTe(d)} \}
\end{equation}
where, $\mathbf{F}_{t}^{SeTe(i)} \in \mathbb{R}^{N \times C}$, $i=\{1,2,...,d\}$, $h=\{1,2,...,H\}$. $\operatorname{Att}$ means the attention operation, $\mathbf{W}_{h}^{Q}, \mathbf{W}_{h}^{K}, \mathbf{W}_{h}^{V} \in \mathbb{R}^{C \times (C/H)}$ project each part of $\mathbf{\hat F}_{t}^{SeTe}$ onto the matrices of queries ($\mathbf{Q}$), keys ($\mathbf{K}$), and values ($\mathbf{V}$) for the $h$-th head, respectively. The outputs of all attention mechanisms are concatenated to get the final result of this module in this layer. The TMA also has a similar structure and definitions as described above.

The TUS is the reverse of the TDS, where the dimension is doubled and the channel is halved. Taking the last TUS as an example, with the concatenated sequence $\mathbf{F}_{t}^{SeTe} \in \mathbb{R}^{C \times N \times d}$, we apply several 1D transposed convolutions to reduce the channel and increase the dimension. At last, the output, $\mathbf{F}_{t}^{TUS}$ has the same size as $\mathbf{F}_t^{MCA}$.

The BEC aims to generate a denoised sample $\mathbf{\hat F}_0$ and estimate the causal direction between pairs of ROIs. As shown in Fig.~\ref{fig4}, the inputs are the noisy samples: $\mathbf{F}_t^{MCA}$ and $\mathbf{F}_t^{TUS}$. After the element adding operation, we can obtain the denoised sample at $t/s$ step:
\begin{equation}
	\mathbf{\hat F}_0 = \mathbf{F}_t^{MCA} + \mathbf{F}_t^{TUS}
\end{equation}
here, we separate the $\mathbf{\hat F}_0$ into multiple rows, where each row represents the corresponding ROI's feature. To mine the causal relationship among ROIs, we introduce the structural equation model (SEM) to predict the direction from one region to another. The causal parameters of SEM can be estimated by:
\begin{equation}
z_{i}' = \sum_{j=1}^{N} \mathbf{A}_{j i} z_{j} + n_i
\end{equation}
where $i=\{1,2,...,N\}$ $\mathbf{A}_{ji}$ indicates the causal effect on $i$-th brain region from $j$-th brain region. $n_i$ is the independent random noise. The matrix $\mathbf{A} \in \mathbb{R}^{N \times N}$ is asymmetric, representing the causal parameters of SEM. The diagonal elements of $\mathbf{A}$ are set to 0 because it is meaningless to consider the effective connectivity of the brain region itself. Therefore, the BEC matrix $\mathbf{A}$ can represent the effective connectivity between any pair of brain regions.

\subsubsection{Multi-resolution consistent Discriminator}
To constrain the generated denoised sample $\mathbf{F}_{t-s}'$ to be consistent with the real sample $\mathbf{F}_{t-s}$ in distribution, we downsample the input sample into four different resolutions and devise four corresponding sub-discriminators to distinguish the synthetic and actual samples. Each discriminator has the conventional transformer structure: layer normalization, self-attention, a feed-forward layer, and the classification head. The output of each discriminator is a scalar in the range of $0 \sim 1$. Averaging all the discriminator's outputs is the final output of the multi-resolution consistent discriminator.

\subsection{Hybrid Loss Functions}
To guarantee the model generates high-quality denoised samples, adversarial loss is introduced to optimize the generator's parameters. There are five kinds of loss functions: the spatial-temporal enhanced generative loss ($L_{SEG}$), the multi-resolution consistent discriminative loss ($L_{MDD}$), the reconstructed loss ($L_{REC}$), the sparse connective penalty loss ($L_{SCP}$), and the classification loss ($L_{CLS}$). We treat the generator as a conditional GAN; when inputting a noisy sample and conditional sample, the generator $\mathbf{G}_{\theta}(\mathbf{F}_t,\mathbf{X},t)$ outputs a synthetic sample $\mathbf{F}_{t-s}'$. And the discriminator $\mathbf{D}_{\theta}$ discriminates whether the sample is from the generator or the forward diffusion. Here are the non-saturating generative and discriminative losses (adversarial diffusive losses):
\begin{equation}
L_{SEG}=\mathbb{E}_{t, q\left(\mathbf{F}_{t} \mid \mathbf{F}_{0}, \mathbf{X}\right), p_{\theta}\left(\mathbf{F}_{t-s} \mid \mathbf{F}_{t}, \mathbf{X}\right)}\left[-\log \left(D_{\theta}\left(\mathbf{F}_{t-s}'\right)\right)\right]
\end{equation}
\begin{equation}
\begin{array}{l}
	\begin{aligned}
	L_{MDD}=&\mathbb{E}_{t, q\left(\mathbf{F}_{t} \mid \mathbf{F}_{0}, \mathbf{X}\right)}\left \{ \mathbb{E}_{q\left(\mathbf{F}_{t-s} \mid \mathbf{F}_{t}, \mathbf{X}\right)}\left[-\log \left(D_{\theta}\left(\mathbf{F}_{t-s}\right)\right)\right]\right. \\
	&+ \mathbb{E}_{p_{\theta}\left(\mathbf{F}_{t-s} \mid \mathbf{F}_{t}, \mathbf{X}\right)}\left[-\log \left(1-D_{\theta}\left(\mathbf{F}_{t-s}'\right)\right)\right]  \}
	\end{aligned}
\end{array}
\end{equation}
after T steps, the final denoised sample $\mathbf{F}_0'$ should recover the clean sample $\mathbf{F}_0$ for every element. The reconstruction loss is defined by:
\begin{equation}
L_{REC} = \mathbb{E}_{ p_{\theta}\left(\mathbf{F}_{0} \mid \mathbf{F}_{s}, \mathbf{X}\right)}[|| \mathbf{F}_0' - \mathbf{F}_0 ||_1]
\end{equation}
%\begin{equation}
%	L_{REC} = \mathbb{E}_{t, q\left(\mathbf{F}_{s}\right), p_{\theta}\left(\mathbf{F}_{0} \mid \mathbf{F}_{s}, \mathbf{X}\right)}[|| \mathbf{F}_0' - \mathbf{F}_0 ||_1]
%\end{equation}
moreover, the sparse effectivity connection between brain regions can be interpreted in brain functional activities. We introduce a penalty on the obtained $\mathbf{A}$ for sparse constraints. Besides, the obtained $\mathbf{A}$ is sent to the classifier $C$ to predict the disease label $y$. The losses are expressed by:
\begin{equation}
L_{SCP} = \gamma \ || \sum_{i=1,j=1}^{N} \mathbf{A}_{i,j} ||
\end{equation}
\begin{equation}
L_{CLS} = \mathbb{E}_{ p_{\theta}\left(\mathbf{A} \mid \mathbf{F}_{s}, \mathbf{X}\right), C(y|\mathbf{A})} (-log(y| \mathbf{A} ))
\end{equation}

\section{Experiments}

\label{s4}
\subsection{Datasets}
In our study, we tested our model on the public Alzheimer's Disease Neuroimaging Initiative (ADNI) dataset\footnote{http://adni.loni.usc.edu/} for classification evaluation. There are 210 subjects scanned with functional magnetic resonance imaging (fMRI), including 61 subjects with late mild cognitive impairment (LMCI), 68 subjects with early mild cognitive impairment (EMCI), and 81 normal controls (NCs). The average ages of LMCI, EMCI, and NC range from 74 to 76. The sex ratio between males and females is nearly balanced. The fMRI is scanned by Siemens with the following scanning parameters: TR = 3.0 s, field strength = 3.0 Tesla, turning angle =80.0 degrees, and the EPI sequence is 197 volumes. There are two ways to preprocess the fMRI. Both of them require the anatomical automatic labeling (AAL90) atlas \cite{tzourio2002automated} for ROI-based time series computing. One is the routine precedence using the GRETNA software to obtain the functional time series, which is treated as the ground truth $\bm F_0$ in the proposed model. The detailed computing steps using GRETNA \cite{wang2015gretna} are described in the work \cite{lei2023multi}. Another one adopted in this paper is using the standard atlas file $aal.nii$ to split each volume of fMRI into 90 brain regions and average all the voxels of each brain region. We discard the first 10 volumes and obtain a matrix with a size of $90 \times 187$, which is the rough sample $\bm X$ input in the model.

% \usepackage{multirow}
\begin{table*}[t!]
	\centering
	\renewcommand\arraystretch{1.4}
	\setlength{\abovecaptionskip}{0pt}%
	\setlength{\belowcaptionskip}{10pt}%
	\caption{Classification performance based on the generated brain networks by different methods (\%).}
	\label{tab1}
	\resizebox{\textwidth}{!}{
		\begin{tabular}{c|l|cccc|cccc|cccc}
			\hline
			\multirow{2}{*}{Method} & \multicolumn{1}{c|}{\multirow{2}{*}{Classifier}} & \multicolumn{4}{c|}{NC vs. EMCI}                                                                                                                                                                                                                                                                  & \multicolumn{4}{c|}{NC vs. LMCI}                                                                                                                                                                                                                                                                  & \multicolumn{4}{c}{EMCI vs. LMCI}                                                                                                                                                                                                                                                                 \\ \cline{3-14}
			& \multicolumn{1}{c|}{}                            & ACC                                                                    & SEN                                                                    & SPE                                                                    & AUC                                                                    & ACC                                                                    & SEN                                                                    & SPE                                                                    & AUC                                                                    & ACC                                                                    & SEN                                                                    & SPE                                                                    & AUC                                                                    \\ \hline
			Empirical               &                                                  & \begin{tabular}[c]{@{}c@{}}69.46\\ ($\pm $1.87)\end{tabular}           & \begin{tabular}[c]{@{}c@{}}70.14\\ ($\pm $2.77)\end{tabular}           & \begin{tabular}[c]{@{}c@{}}68.88\\ ($\pm $ 3.28)\end{tabular}          & \begin{tabular}[c]{@{}c@{}}69.98\\ ($\pm $ 3.48)\end{tabular}          & \begin{tabular}[c]{@{}c@{}}76.26\\ ($\pm $ 1.81)\end{tabular}          & \begin{tabular}[c]{@{}c@{}}74.91\\ ($\pm $ 2.79)\end{tabular}          & \begin{tabular}[c]{@{}c@{}}77.28\\ ($\pm $ 2.03)\end{tabular}          & \begin{tabular}[c]{@{}c@{}}76.69\\ ($\pm $ 3.45)\end{tabular}          & \begin{tabular}[c]{@{}c@{}}76.66\\ ($\pm $ 1.65)\end{tabular}          & \begin{tabular}[c]{@{}c@{}}76.72\\ ($\pm $ 1.69)\end{tabular}          & \begin{tabular}[c]{@{}c@{}}76.61\\ ($\pm $ 2.13)\end{tabular}          & \begin{tabular}[c]{@{}c@{}}77.99\\ ($\pm $ 2.95)\end{tabular}          \\ \cline{3-14}
			GCCA                    &                                                  & \begin{tabular}[c]{@{}c@{}}75.30\\ ($\pm $ 1.57)\end{tabular}          & \begin{tabular}[c]{@{}c@{}}75.14\\ ($\pm $ 2.13)\end{tabular}          & \begin{tabular}[c]{@{}c@{}}75.43\\ ($\pm $ 2.13)\end{tabular}          & \begin{tabular}[c]{@{}c@{}}76.20\\ ($\pm $ 2.48)\end{tabular}          & \begin{tabular}[c]{@{}c@{}}83.16\\ ($\pm $ 1.60)\end{tabular}          & \begin{tabular}[c]{@{}c@{}}82.62\\ ($\pm $ 2.21)\end{tabular}          & \begin{tabular}[c]{@{}c@{}}83.58\\ ($\pm $ 1.93)\end{tabular}          & \begin{tabular}[c]{@{}c@{}}83.03\\ ($\pm $ 1.83)\end{tabular}          & \begin{tabular}[c]{@{}c@{}}82.48\\ ($\pm $ 1.42)\end{tabular}          & \begin{tabular}[c]{@{}c@{}}81.97\\ ($\pm $ 1.89)\end{tabular}          & \begin{tabular}[c]{@{}c@{}}82.93\\ ($\pm $ 1.85)\end{tabular}          & \begin{tabular}[c]{@{}c@{}}82.98\\ ($\pm $ 2.72)\end{tabular}          \\ \cline{3-14}
			EC-GAN                  & \multicolumn{1}{c|}{SVM}                         & \begin{tabular}[c]{@{}c@{}}83.08\\ ($\pm $ 1.17)\end{tabular}          & \begin{tabular}[c]{@{}c@{}}82.93\\ ($\pm $ 1.85)\end{tabular}          & \begin{tabular}[c]{@{}c@{}}83.21\\ ($\pm $ 2.11)\end{tabular}          & \begin{tabular}[c]{@{}c@{}}82.69\\ ($\pm $ 2.23)\end{tabular}          & \begin{tabular}[c]{@{}c@{}}90.35\\ ($\pm $ 1.19)\end{tabular}          & \begin{tabular}[c]{@{}c@{}}90.32\\ ($\pm $ 1.21)\end{tabular}          & \begin{tabular}[c]{@{}c@{}}90.37\\ ($\pm $ 2.08)\end{tabular}          & \begin{tabular}[c]{@{}c@{}}90.53\\ ($\pm $ 1.33)\end{tabular}          & \begin{tabular}[c]{@{}c@{}}87.59\\ ($\pm $ 1.21)\end{tabular}          & \begin{tabular}[c]{@{}c@{}}87.21\\ ($\pm $ 1.85)\end{tabular}          & \begin{tabular}[c]{@{}c@{}}87.94\\ ($\pm $ 1.81)\end{tabular}          & \begin{tabular}[c]{@{}c@{}}86.65\\ ($\pm $ 1.83)\end{tabular}          \\ \cline{3-14}
			STGCM                   &                                                  & \begin{tabular}[c]{@{}c@{}}84.09\\ ($\pm $ 0.95)\end{tabular}          & \begin{tabular}[c]{@{}c@{}}85.14\\ ($\pm $ 0.83)\end{tabular}          & \begin{tabular}[c]{@{}c@{}}83.21\\ ($\pm $ 1.66)\end{tabular}          & \begin{tabular}[c]{@{}c@{}}85.15\\ ($\pm $ 1.63)\end{tabular}          & \begin{tabular}[c]{@{}c@{}}92.25\\ ($\pm $ 1.04)\end{tabular}          & \begin{tabular}[c]{@{}c@{}}92.12\\ ($\pm $ 1.50)\end{tabular}          & \begin{tabular}[c]{@{}c@{}}92.34\\ ($\pm $ 1.51)\end{tabular}          & \begin{tabular}[c]{@{}c@{}}92.50\\ ($\pm $ 1.12)\end{tabular}          & \begin{tabular}[c]{@{}c@{}}90.54\\ ($\pm $ 1.40)\end{tabular}          & \begin{tabular}[c]{@{}c@{}}91.47\\ ($\pm $ 2.01)\end{tabular}          & \begin{tabular}[c]{@{}c@{}}89.71\\ ($\pm $ 0.98)\end{tabular}          & \begin{tabular}[c]{@{}c@{}}89.67\\ ($\pm $ 1.83)\end{tabular}          \\ \cline{3-14}
			Ours                    &                                                  & \textbf{\begin{tabular}[c]{@{}c@{}}85.43\\ ($\pm $ 0.78)\end{tabular}} & \textbf{\begin{tabular}[c]{@{}c@{}}86.31\\ ($\pm $ 0.99)\end{tabular}} & \textbf{\begin{tabular}[c]{@{}c@{}}84.69\\ ($\pm $ 0.86)\end{tabular}} & \textbf{\begin{tabular}[c]{@{}c@{}}86.91\\ ($\pm $ 1.03)\end{tabular}} & \textbf{\begin{tabular}[c]{@{}c@{}}93.45\\ ($\pm $ 0.88)\end{tabular}} & \textbf{\begin{tabular}[c]{@{}c@{}}93.93\\ ($\pm $ 1.10)\end{tabular}} & \textbf{\begin{tabular}[c]{@{}c@{}}93.08\\ ($\pm $ 1.04)\end{tabular}} & \textbf{\begin{tabular}[c]{@{}c@{}}93.68\\ ($\pm $ 1.13)\end{tabular}} & \textbf{\begin{tabular}[c]{@{}c@{}}91.70\\ ($\pm $ 0.97)\end{tabular}} & \textbf{\begin{tabular}[c]{@{}c@{}}91.63\\ ($\pm $ 1.21)\end{tabular}} & \textbf{\begin{tabular}[c]{@{}c@{}}91.76\\ ($\pm $ 1.23)\end{tabular}} & \textbf{\begin{tabular}[c]{@{}c@{}}91.65\\ ($\pm $ 1.26)\end{tabular}} \\ \hline
			Empirical               &                                                  & \begin{tabular}[c]{@{}c@{}}70.93\\ ($\pm $ 1.34)\end{tabular}          & \begin{tabular}[c]{@{}c@{}}70.73\\ ($\pm $ 2.01)\end{tabular}          & \begin{tabular}[c]{@{}c@{}}71.11\\ ($\pm $ 1.76)\end{tabular}          & \begin{tabular}[c]{@{}c@{}}71.87\\ ($\pm $ 1.83)\end{tabular}          & \begin{tabular}[c]{@{}c@{}}78.87\\ ($\pm $ 1.62)\end{tabular}          & \begin{tabular}[c]{@{}c@{}}78.19\\ ($\pm $ 2.79)\end{tabular}          & \begin{tabular}[c]{@{}c@{}}79.38\\ ($\pm $ 2.02)\end{tabular}          & \begin{tabular}[c]{@{}c@{}}78.20\\ ($\pm $ 2.17)\end{tabular}          & \begin{tabular}[c]{@{}c@{}}77.51\\ ($\pm $ 1.71)\end{tabular}          & \begin{tabular}[c]{@{}c@{}}78.03\\ ($\pm $ 1.58)\end{tabular}          & \begin{tabular}[c]{@{}c@{}}77.05\\ ($\pm $ 2.70)\end{tabular}          & \begin{tabular}[c]{@{}c@{}}78.89\\ ($\pm $ 2.71)\end{tabular}          \\ \cline{3-14}
			GCCA                    &                                                  & \begin{tabular}[c]{@{}c@{}}76.24\\ ($\pm $ 1.38)\end{tabular}          & \begin{tabular}[c]{@{}c@{}}76.61\\ ($\pm $ 2.54)\end{tabular}          & \begin{tabular}[c]{@{}c@{}}75.92\\ ($\pm $ 1.86)\end{tabular}          & \begin{tabular}[c]{@{}c@{}}77.48\\ ($\pm $ 2.62)\end{tabular}          & \begin{tabular}[c]{@{}c@{}}83.73\\ ($\pm $ 1.42)\end{tabular}          & \begin{tabular}[c]{@{}c@{}}83.77\\ ($\pm $ 1.80)\end{tabular}          & \begin{tabular}[c]{@{}c@{}}83.70\\ ($\pm $ 1.91)\end{tabular}          & \begin{tabular}[c]{@{}c@{}}84.13\\ ($\pm $ 1.77)\end{tabular}          & \begin{tabular}[c]{@{}c@{}}82.63\\ ($\pm $ 1.22)\end{tabular}          & \begin{tabular}[c]{@{}c@{}}82.95\\ ($\pm $ 1.92)\end{tabular}          & \begin{tabular}[c]{@{}c@{}}82.35\\ ($\pm $ 1.69)\end{tabular}          & \begin{tabular}[c]{@{}c@{}}82.45\\ ($\pm $ 2.64)\end{tabular}          \\ \cline{3-14}
			EC-GAN                  & \multicolumn{1}{c|}{BrainnetCNN}                 & \begin{tabular}[c]{@{}c@{}}83.69\\ ($\pm $ 1.09)\end{tabular}          & \begin{tabular}[c]{@{}c@{}}83.08\\ ($\pm $ 1.86)\end{tabular}          & \begin{tabular}[c]{@{}c@{}}84.20\\ ($\pm $ 1.27)\end{tabular}          & \begin{tabular}[c]{@{}c@{}}84.35\\ ($\pm $ 1.41)\end{tabular}          & \begin{tabular}[c]{@{}c@{}}90.56\\ ($\pm $ 0.94)\end{tabular}          & \begin{tabular}[c]{@{}c@{}}90.48\\ ($\pm $ 1.29)\end{tabular}          & \begin{tabular}[c]{@{}c@{}}90.61\\ ($\pm $ 1.19)\end{tabular}          & \begin{tabular}[c]{@{}c@{}}90.24\\ ($\pm $ 1.53)\end{tabular}          & \begin{tabular}[c]{@{}c@{}}88.45\\ ($\pm $ 1.28)\end{tabular}          & \begin{tabular}[c]{@{}c@{}}88.03\\ ($\pm $ 1.73)\end{tabular}          & \begin{tabular}[c]{@{}c@{}}88.82\\ ($\pm $ 1.42)\end{tabular}          & \begin{tabular}[c]{@{}c@{}}89.71\\ ($\pm $ 1.71)\end{tabular}          \\ \cline{3-14}
			STGCM                   &                                                  & \begin{tabular}[c]{@{}c@{}}85.09\\ ($\pm $ 0.88)\end{tabular}          & \begin{tabular}[c]{@{}c@{}}84.85\\ ($\pm $ 1.55)\end{tabular}          & \begin{tabular}[c]{@{}c@{}}85.31\\ ($\pm $ 1.58)\end{tabular}          & \begin{tabular}[c]{@{}c@{}}85.41\\ ($\pm $ 1.15)\end{tabular}          & \begin{tabular}[c]{@{}c@{}}93.52\\ ($\pm $ 0.98)\end{tabular}          & \begin{tabular}[c]{@{}c@{}}94.26\\ ($\pm $ 1.39)\end{tabular}          & \begin{tabular}[c]{@{}c@{}}92.96\\ ($\pm $ 1.17)\end{tabular}          & \begin{tabular}[c]{@{}c@{}}93.06\\ ($\pm $ 1.56)\end{tabular}          & \begin{tabular}[c]{@{}c@{}}91.47\\ ($\pm $ 0.89)\end{tabular}          & \begin{tabular}[c]{@{}c@{}}92.62\\ ($\pm $ 1.15)\end{tabular}          & \begin{tabular}[c]{@{}c@{}}90.44\\ ($\pm $ 1.24)\end{tabular}          & \begin{tabular}[c]{@{}c@{}}90.76\\ ($\pm $ 1.47)\end{tabular}          \\ \cline{3-14}
			Ours                    &                                                  & \textbf{\begin{tabular}[c]{@{}c@{}}86.58\\ ($\pm $ 0.44)\end{tabular}} & \textbf{\begin{tabular}[c]{@{}c@{}}86.02\\ ($\pm $ 0.77)\end{tabular}} & \textbf{\begin{tabular}[c]{@{}c@{}}87.03\\ ($\pm $ 0.87)\end{tabular}} & \textbf{\begin{tabular}[c]{@{}c@{}}87.11\\ ($\pm $ 0.86)\end{tabular}} & \textbf{\begin{tabular}[c]{@{}c@{}}94.57\\ ($\pm $ 0.74)\end{tabular}} & \textbf{\begin{tabular}[c]{@{}c@{}}94.75\\ ($\pm $ 1.29)\end{tabular}} & \textbf{\begin{tabular}[c]{@{}c@{}}94.44\\ ($\pm $ 0.64)\end{tabular}} & \textbf{\begin{tabular}[c]{@{}c@{}}94.49\\ ($\pm $ 1.16)\end{tabular}} & \textbf{\begin{tabular}[c]{@{}c@{}}92.24\\ ($\pm $ 0.63)\end{tabular}} & \begin{tabular}[c]{@{}c@{}}92.45\\ ($\pm $ 0.84)\end{tabular}          & \textbf{\begin{tabular}[c]{@{}c@{}}92.06\\ ($\pm $ 0.75)\end{tabular}} & \textbf{\begin{tabular}[c]{@{}c@{}}92.45\\ ($\pm $ 0.58)\end{tabular}} \\ \hline
	\end{tabular}}
\end{table*}

\subsection{Training Settings and Evaluation Metrics}
In the training process, the input of our model is the 4D functional MRI, and the output is the ROI-based time series and the BECs. We set the parameters as follows: $T$ = 1000, $s$ = 250. $C$ = 2. $N$ = 90, $d$ = 187, $L_i$ = 2 ($i=0,1,2,3,4$), $\gamma=1.9$. The Pytorch framework is used to optimize the model's weightings under an Ubuntu 18.04 system. The batch size is 16, and the total epochs are 600. The learning rates for the generator and discriminator are 0.001 and 0.0002, respectively.

We adopt 5-fold cross-validation in our model's validation. Specifically, the subjects in each category are randomly divided into five parts. The model is trained on the four parts of them and tested on the rest. The final accuracy is computed by averaging the results from the five parts. After obtaining the BECs, we conduct three binary classification tasks (i.e., NC vs. EMCI, NC vs. LMCI, and EMCI vs. LMCI) for the model's performance evaluation. Two commonly used classifiers are adopted to evaluate the classification performance, including the support vector machine (SVM) and the BrainNetCNN\cite{kawahara2017brainnetcnn}. The evaluation metrics are the area under the receiver operating characteristic curve (AUC), the prediction accuracy (ACC), the positive sensitivity (SEN), and the negative specificity (SPE).

\subsection{Prediction Results}
In order to show the superior performance of the generated BECs, we introduce four other methods to compare the classification performance. (1) the empirical method; (2) the Granger causal connectivity analysis (GCCA)\cite{seth2010matlab}; (3) the effective connectivity based on generative adversarial networks (EC-GAN)\cite{liu2020ec}; (4) the spatiotemporal graph convolutional models (STGCM)\cite{zou2022exploring}.
We repeated the classification experiment 10 times. For each repetition, we randomly divided the dataset into five parts and adopted the 5-fold cross-validation strategy to train the model. The final mean and standard deviation values are computed from the results of 10 repetition experiments.
The classification results are shown in Table~\ref{tab1}. Compared with the empirical method, the other four methods achieve better classification performance in both classifiers by generating BECs. This indicates that effective connectivity contains the causal information that is correlated with MCI. Among the four BEC-based methods, our model achieves the best mean values for ACC, SEN, SPE, and AUC with 86.58\%, 86.02\%, 87.03\%, and 87.11\% for NC vs. EMCI, respectively. The best mean value of ACC, SEN, SPE, and AUC for the NC vs. LMCI task are 94.57\%, 94.75\%, 94.44\%, and 94.49\%. The values of 92.24\%, 92.45\%, 92.06\%, and 92.45\% are obtained using our model in the EMC vs. LMCI prediction task. The results demonstrate that the generated effective connectivity (our model) gives better classification performance than the functional connectivity (empirical method), indicating its greater ability to capture MCI-related information.

\begin{figure*}[t!]
	\centering
	\includegraphics[width=0.8\textwidth]{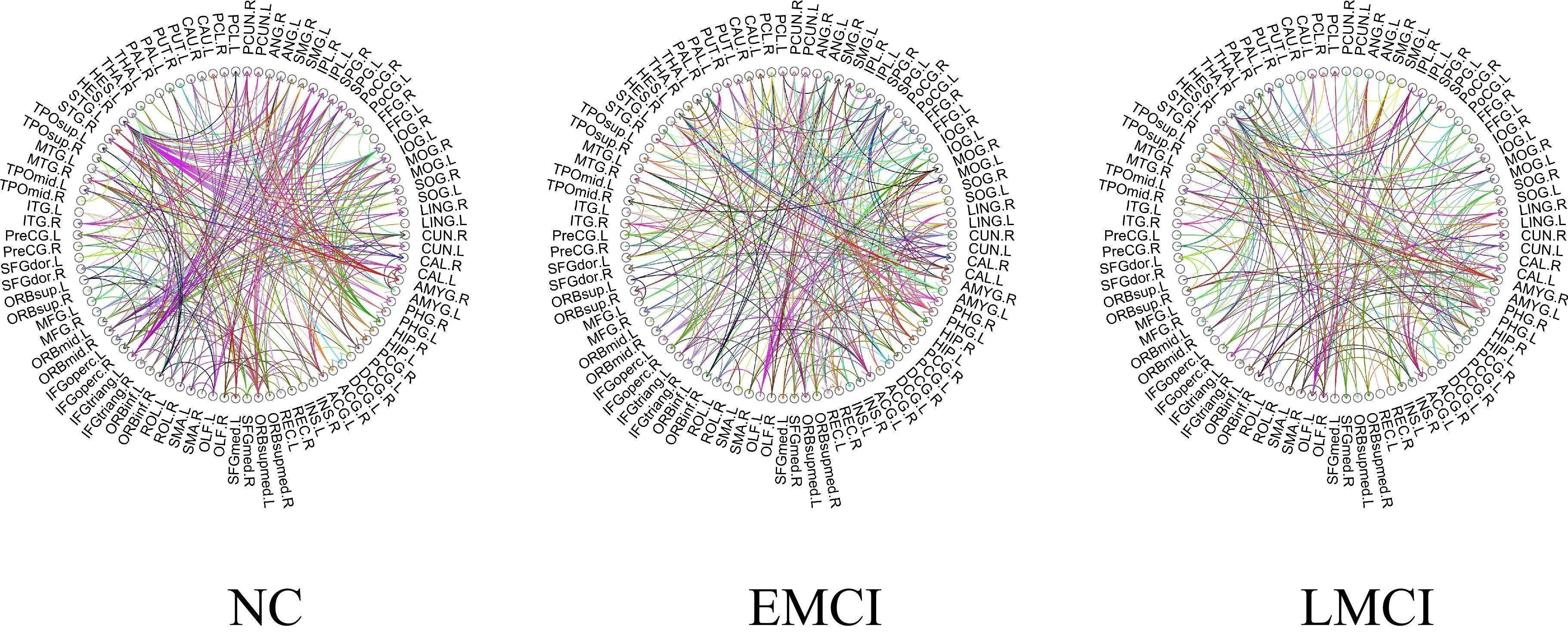}
	\caption{The generated brain effective connectivities at NC, EMCI, and LMCI, respectively. The arcs with arrows are the effective connection, and the colors have no meaning. The circles on the outside represent the brain regions. \label{fig7}}
\end{figure*}

\subsection{Effective Connectivity Analysis}
\begin{figure*}[htbp]
	\centering
	\includegraphics[width=0.8\textwidth]{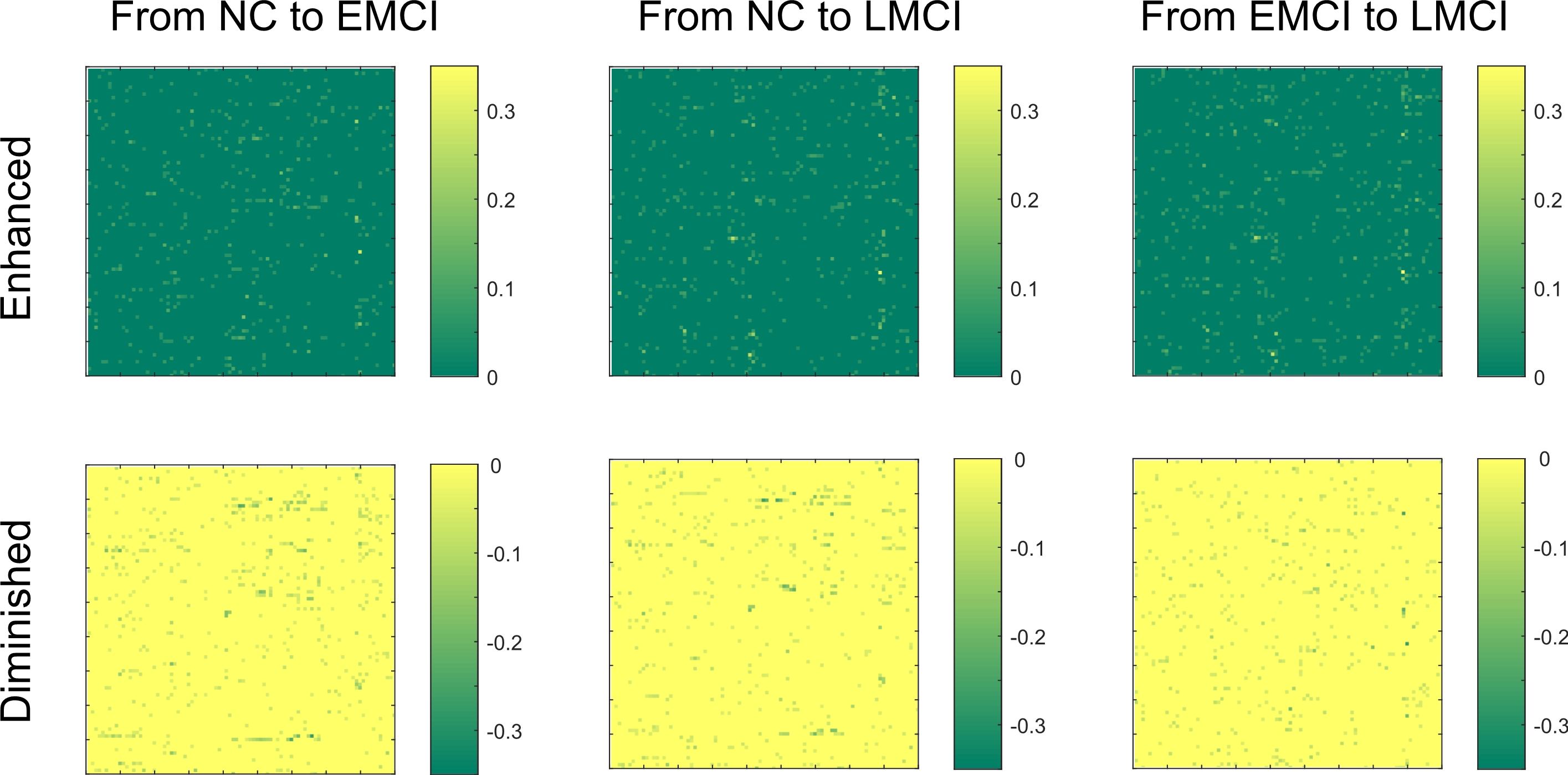}
	\caption{The altered brain effective connectivities from NC to EMCI, from NC to LMCI, and from EMCI to LMCI, respectively. The top row represents the enhanced connections, and the bottom row represents the diminished connections. \label{fig8}}
\end{figure*}

In addition to the fact that brain regions play an important role in disease diagnosis, the causal relationship between them can uncover the underlying pathological mechanism of MCI. In this section, we analyze the generated BECs and predict the abnormal directional connections for further study. To make the results statistically significant, we average all the BECs for each category and filter out the values that fall below the threshold of 0.1. The averaged BEC at three different stages is shown in Fig.~\ref{fig7} by modifying the circularGraph packages\footnote{https://github.com/paul-kassebaum-mathworks/circularGraph}. We compute the altered effective connectivity by subtracting the averaged BEC matrix at the early stage from the later stage. As a result, a total of six altered effective connectivity matrices are obtained, consisting of the enhanced and diminished connectivities from NC to EMCI, from NC to LMCI, and from EMCI to LMCI. The altered effective connectivities are shown in Fig.~\ref{fig8}. The top row represents the enhanced connections, and the bottom row represents the diminished connections. Each matrix is asymmetric, and the element values range from $-0.35 \sim 0.35$. The positive value means the directional connection strength is enhanced, while the negative value means the directional connection strength is diminished.

\begin{figure}[htbp]
	\centering
	\includegraphics[width=0.7\columnwidth]{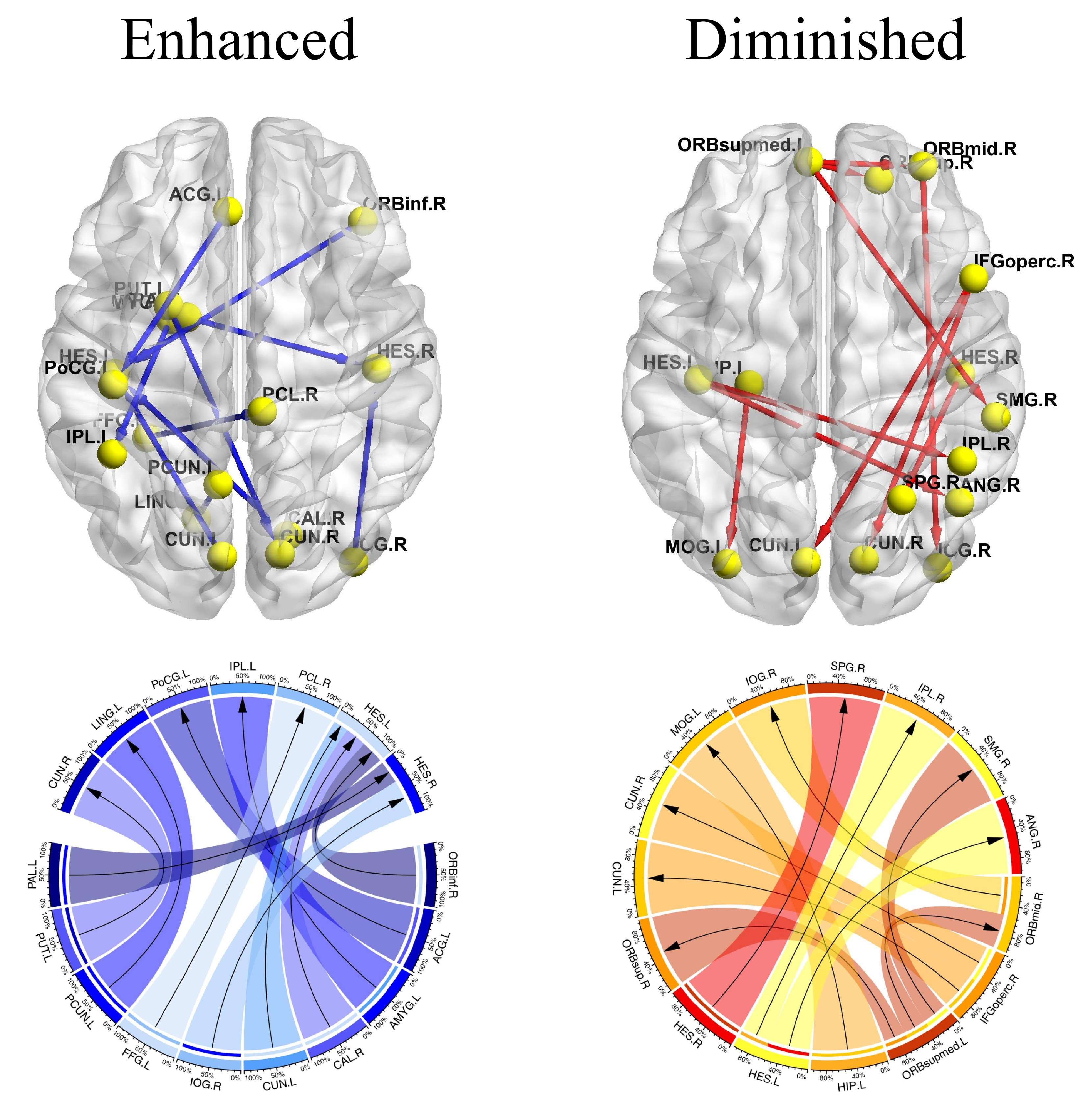}
	\caption{The top 10 enhanced and diminished effective connections from NC to EMCI. \label{fig9}}
\end{figure}

\begin{figure}[htbp]
	\centering
	\includegraphics[width=0.7\columnwidth]{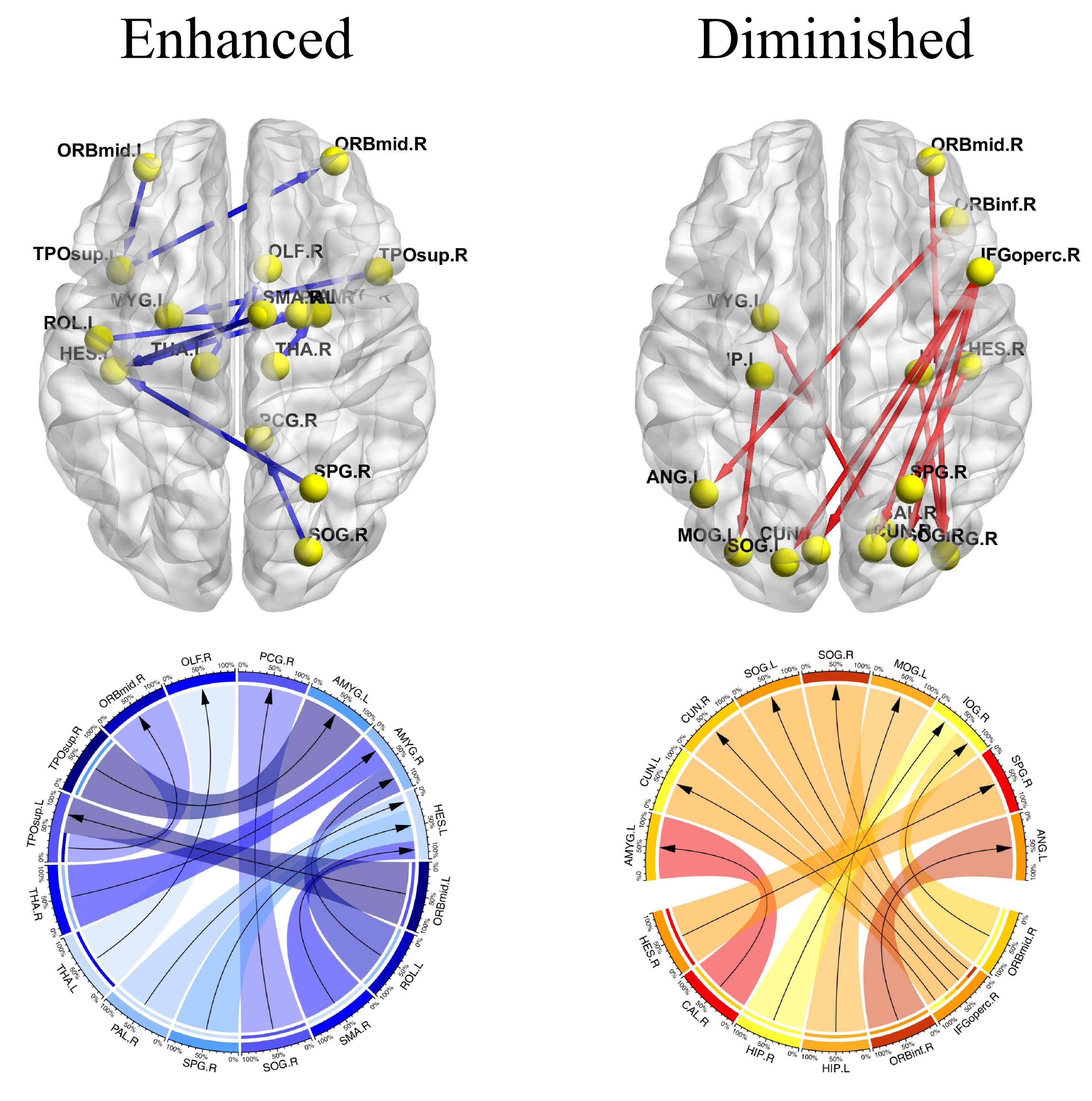}
	\caption{The top 10 enhanced and diminished effective connections from NC to LMCI. \label{fig10}}
\end{figure}

\begin{figure}[htbp]
	\centering
	\includegraphics[width=0.7\columnwidth]{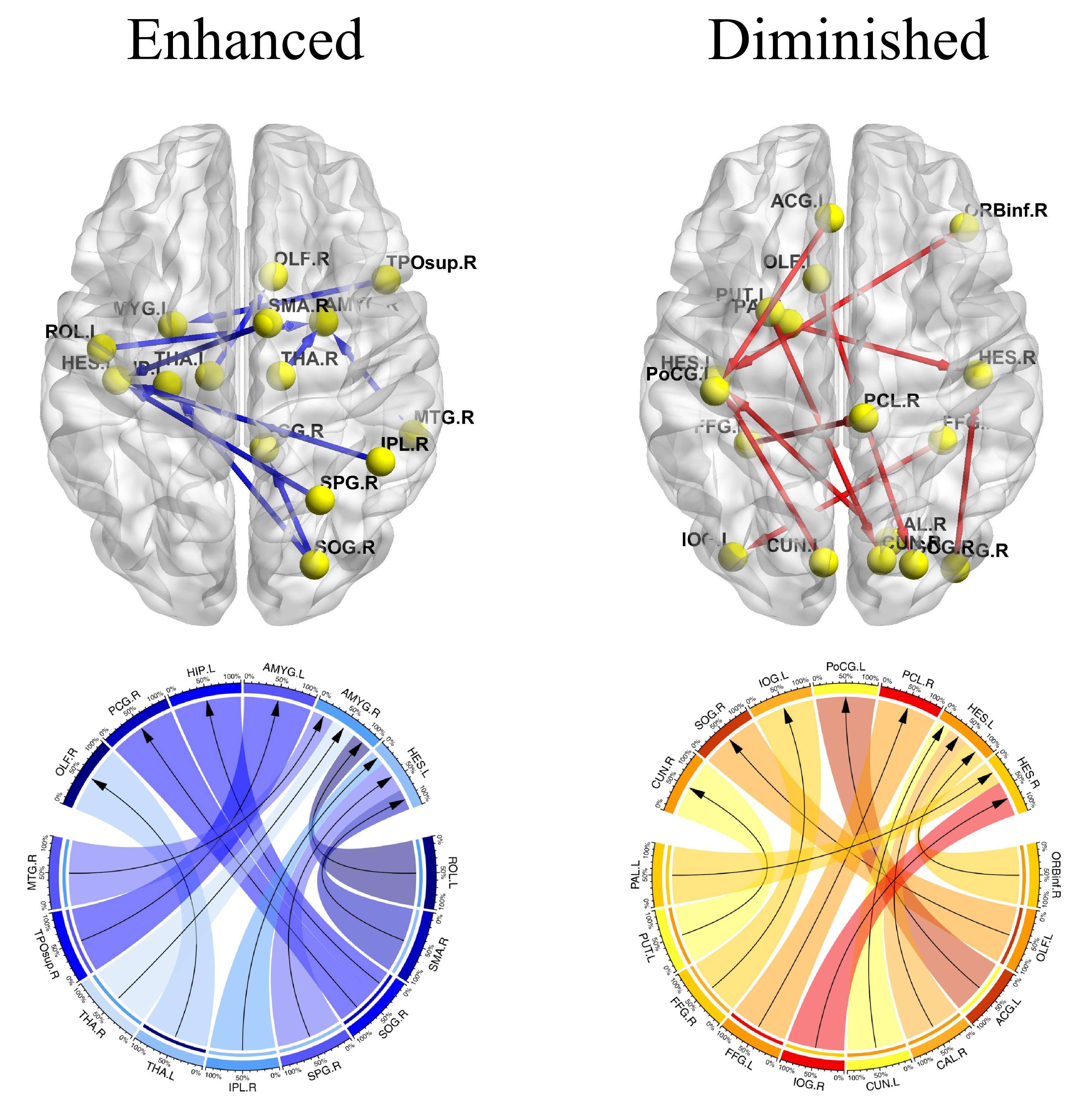}
	\caption{The top 10 enhanced and diminished effective connections from EMCI to LMCI. \label{fig11}}
\end{figure}

These altered effective connections probably contribute to the cause of MCI. To find the important effective connections during the MCI progression, we sort the altered effective connections and calculate the top 10 enhanced and diminished effective connections. The results are shown in Fig.~\ref{fig9}, Fig.~\ref{fig10}, and Fig.~\ref{fig11}.
The enhanced and diminished connections are displayed on both the brain atlas view and the circularGraph view. From NC to EMCI, the 10 effective connections with the greatest enhancement in connection strength are
ORBinf.R $\rightarrow$ HES.L, ACG.L $\rightarrow$ PoCG.L, AMYG.L $\rightarrow$ IPL.L, CAL.R $\rightarrow$ HES.L, CUN.L $\rightarrow$ HES.L, IOG.R $\rightarrow$ HES.R, FFG.L $\rightarrow$ PCL.R, PCUN.L $\rightarrow$ LING.L, PUT.L $\rightarrow$ CUN.R, PAL.L $\rightarrow$ HES.R;
The 10 effective connections with the greatest diminishment in connection strength are
ORBmid.R $\rightarrow$ IOG.R, IFGoperc.R $\rightarrow$ CUN.L, IFGoperc.R $\rightarrow$ CUN.R, ORBsupmed.L $\rightarrow$ ORBsup.R, ORBsupmed.L $\rightarrow$ ORBmid.R, ORBsupmed.L $\rightarrow$ SMG.R, HIP.L $\rightarrow$ MOG.L, HES.L $\rightarrow$ IPL.R, HES.L $\rightarrow$ ANG.R, HES.R $\rightarrow$ SPG.R.
As the EMCI progresses to the LMCI, the top 10 enhanced effective connections are:
ROL.L $\rightarrow$ AMYG.R, SMA.R $\rightarrow$ HES.L, SOG.R $\rightarrow$ PCG.R, SOG.R $\rightarrow$ HIP.L, SPG.R $\rightarrow$ HES.L, IPL.R $\rightarrow$ HES.L, THA.L $\rightarrow$ OLF.R, THA.R $\rightarrow$ AMYG.R, TPOsup.R $\rightarrow$ AMYG.L, MTG.R $\rightarrow$ AMYG.R;
and the top 10 diminished effective connections are
ORBinf.R $\rightarrow$ HES.L, OLF.L $\rightarrow$ SOG.R, ACG.L $\rightarrow$ PoCG.L, CAL.R $\rightarrow$ HES.L, CUN.L $\rightarrow$ HES.L, IOG.R $\rightarrow$ HES.R, FFG.L $\rightarrow$ PCL.R, FFG.R $\rightarrow$ IOG.L, PUT.L $\rightarrow$ CUN.R, PAL.L $\rightarrow$ HES.R.
The enhanced effective connections between NC and LMCI groups includes
ORBmid.L $\rightarrow$ TPOsup.L, ROL.L $\rightarrow$ AMYG.R, SMA.R $\rightarrow$ HES.L, SOG.R $\rightarrow$ PCG.R, SPG.R $\rightarrow$ HES.L, PAL.R $\rightarrow$ HES.L, THA.L $\rightarrow$ OLF.R, THA.R $\rightarrow$ AMYG.R, TPOsup.L $\rightarrow$ ORBmid.R, TPOsup.R $\rightarrow$ AMYG.L;
and the diminished effective connections are
ORBmid.R $\rightarrow$ IOG.R, IFGoperc.R $\rightarrow$ CUN.L, IFGoperc.R $\rightarrow$ CUN.R, IFGoperc.R $\rightarrow$ SOG.L, IFGoperc.R $\rightarrow$ SOG.R, ORBinf.R $\rightarrow$ ANG.L, HIP.L $\rightarrow$ MOG.L, HIP.R $\rightarrow$ IOG.R, CAL.R $\rightarrow$ AMYG.L, HES.R $\rightarrow$ SPG.R.
These connection-related ROIs are consistent with the top nine important ROIs identified in the above section.


\subsection{Ablation Study}
\begin{figure}[htbp]
	\centering
	\includegraphics[width=\columnwidth]{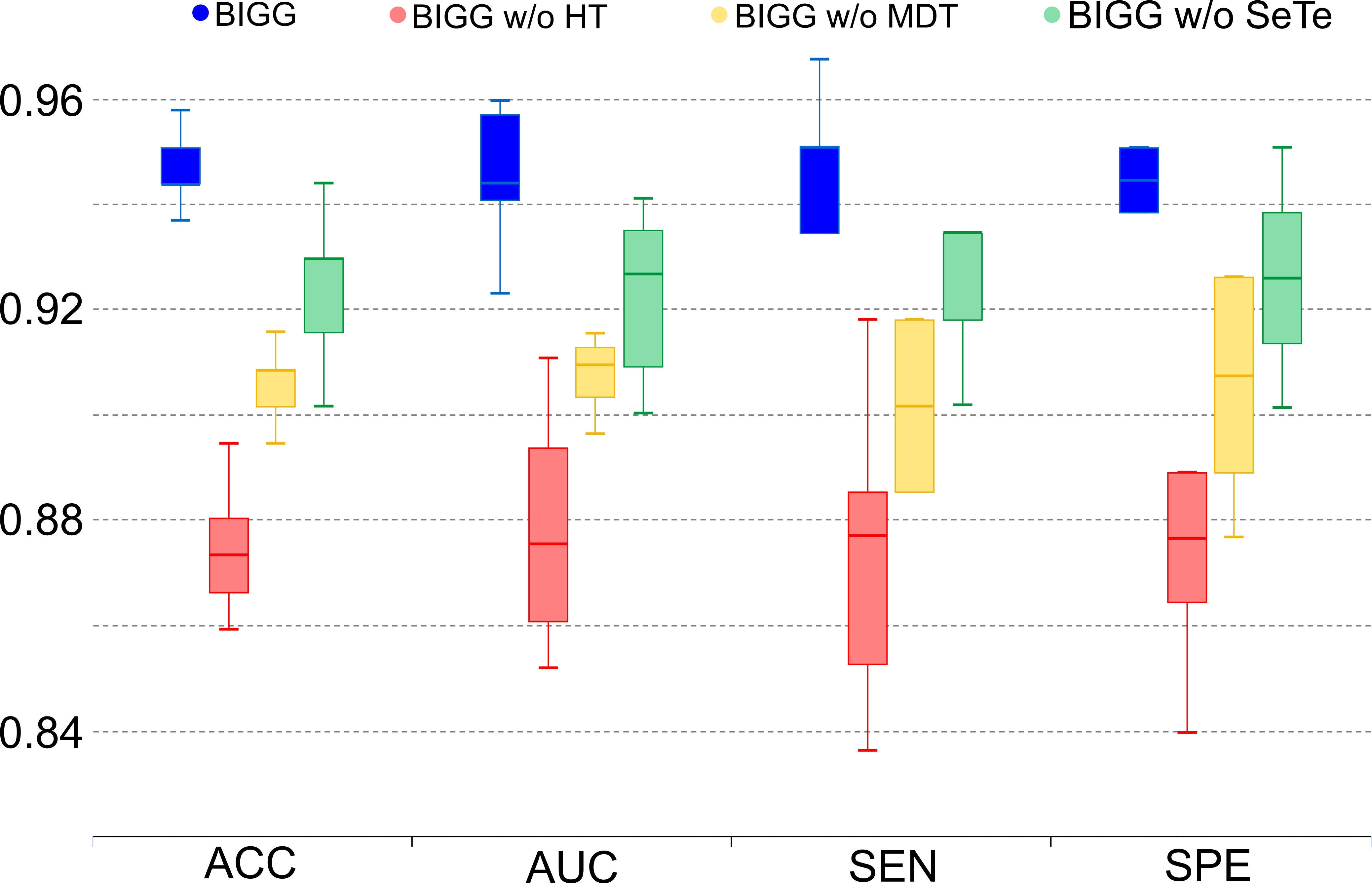}
	\caption{The impact of the proposed generator and discriminator on the model's performance. \label{fig12}}
\end{figure}

In our experiment, the BEC is obtained by optimizing the generator and the discriminator. To investigate whether the proposed generator and discriminator are effective, we design three variants of the proposed model and repeat ten times the 5-fold cross-validations. (1) BIGG without hierarchical transformer (BIGG w/o HT). We removed the TDS and TUS and only kept one SeTe block in the generator. (2) BIGG without SeTe blocks (BIGG w/o SeTe). In this case, we replace the SeTe with conventional 1D convolution with a kerner size of $1\times3$. (3) BIGG without multiresolution diffusive transformer (BIGG w/o MDT). We remove the downsampling operation in the discriminator and keep the $D_1$. For each variant, we compute the ACCs, AUCs, SENs, and SPEs. The results are shown in Fig.~\ref{fig12}. It can be observed that removing the hierarchical structure greatly reduces the classification performance, which shows the effectiveness and necessity of the hierarchical structure for BEC generation. The SeTe block and the discriminator structure also lower the model's classification performance to some extent.

\begin{table}[]
	\renewcommand\arraystretch{1.4}
	\setlength{\abovecaptionskip}{0pt}%
	\setlength{\belowcaptionskip}{10pt}%
	\caption{The top twelve important effective connections that are highly correlated with MCI. + means enhanced connection, and - means diminished connection.}\label{tab2}
	\resizebox{\columnwidth}{!}{\begin{tabular}{ccc}
		\hline
		Type               & Effective connection            & Location                        \\ \hline
		\multirow{6}{*}{+}       & ORBmid.L$\rightarrow$TPOsup.L          & Frontal lobe/Temporal lobe      \\
		& SOG.R$\rightarrow$PCG.R               & Occipital lobe/Parietal lobe  \\
		& PAL.R$\rightarrow$HES.L                & Subcortical areas/Temporal lobe \\
		& THA.L$\rightarrow$OLF.R                & Subcortical areas/Frontal lobe    \\
		& THA.R$\rightarrow$AMYG.R               & Subcortical areas/Temporal lobe \\
		& TPOsup.L$\rightarrow$ORBmid.R          & Temporal lobe/Frontal lobe      \\ \hline
		\multirow{6}{*}{-}       & IFGoperc.R$\rightarrow$SOG.L           & Frontal lobe/Occipital lobe     \\
		& HIP.L$\rightarrow$MOG.L                & Temporal lobe/Occipital lobe    \\
		& HIP.L$\rightarrow$MOG.R                & Temporal lobe/Occipital lobe    \\
		& HIP.R$\rightarrow$IOG.R                & Temporal lobe/Occipital lobe    \\
		& CAL.R$\rightarrow$AMYG.L               & Occipital lobe/Temporal lobe    \\
		& HES.L$\rightarrow$ANG.L                & Temporal lobe/Parietal lobe     \\ \hline
	\end{tabular}}
\end{table}

\begin{figure*}[htbp]
	\centering
	\includegraphics[width=0.8\textwidth]{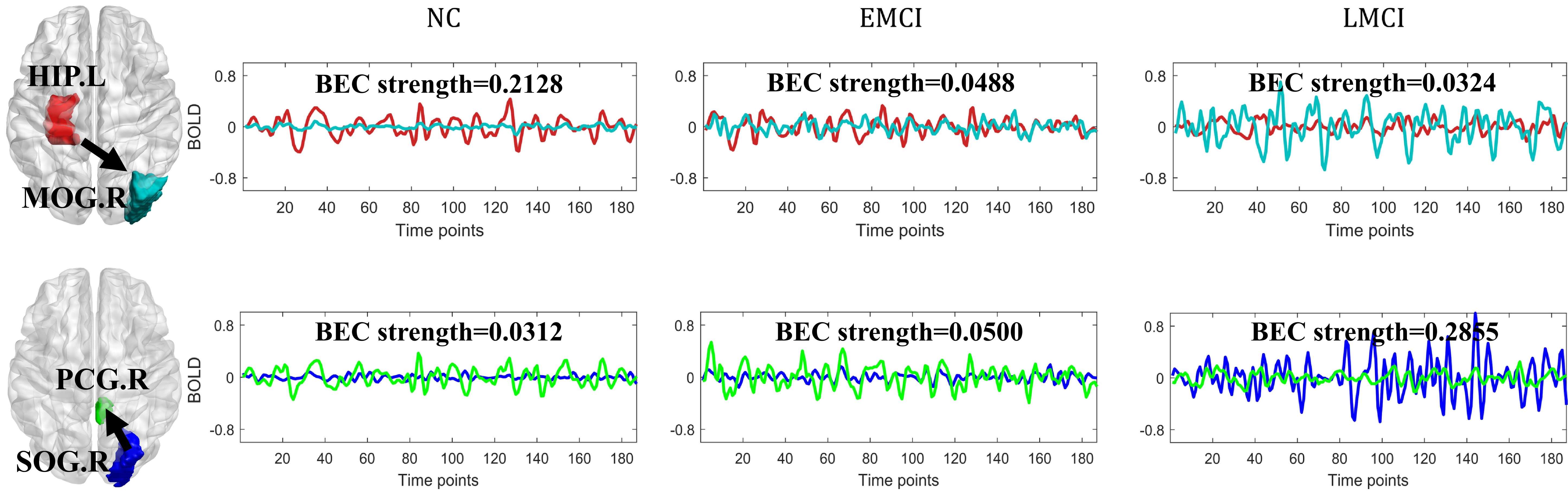}
	\caption{Visualization of ROI-based BOLD signals from two effective connections. The red arrow is the causal direction. The top row is the diminished connection, and the bottom row is the enhanced connection. \label{fig14}}
\end{figure*}

\section{Discussion}
\label{s5}
The proposed model can generate BEC from 4D fMRI in an end-to-end manner for MCI analysis. The ROI mask of the AAL90 atlas helps to parcellate the whole 3D volume into 90 ROIs at each time point. This operation contains many linear interpolations and ignores voxels at the boundaries of adjacent brain regions, which results in a rough ROI-based time series. To denoise the rough ROI-based time-series obtained, we employ the conditional DDPM to successively remove the noise and get a clean ROI-based time-series because of its powerful ability to generate high-quality and diverse results. As the denoising process needs thousands of steps to approach clean data, we introduce the adversarial strategy to speed the denoising process. As displayed in Fig.~\ref{fig12}, when removing the adversarial strategy (BIGG w/o MDT), the classification performance shows a significant decrease. Also, the hierarchical structure and the SeTe block in the generator both ensure good generation results because they focus on the multi-scale spatiotemporal features and thus enhance the denoising quality.

The altered effective connections detected are important for the three scenarios. These connections are partly correlated, which is essential for discovering MCI-related biomarkers. We focus on the same altered connections between NC vs. EMCI and NC vs. LMCI. The top 12 MCI-related effective connections are shown in Table.~\ref{tab2}. The enhanced and diminished effective connection-related ROIs are identified in previous studies\cite{lin2019multiparametric,lei2020self,chen2021abnormal}.
For example, the HIP has characteristics of decreasing volume and diminishing connection strength as the disease progresses\cite{schuff2009mri,tahmasian2015lower}. Also, the AMYG has been reported to process both sensory information and punishment/reward-related learning memory \cite{yang2023plastic}. Disruption of AMYG-related connections can bring cognitive decline\cite{ortner2016progressively}. The ANG can correlate visual information with language expression. Patients with MCI lose ANG-related connections and cannot read the visual signals\cite{lee2016default}.
We display two examples for visualizing the effective connection-strength-changing process. As shown in Fig.~\ref{fig14}, the effective connection from SOG.R to PCG.R is becoming weaker as NC progresses to LMCI. This perhaps results in the memory problem and is consistent with clinical works\cite{xue2019altered,berron2020medial}. The effective connection from HIP.L to MOG.R becomes progressively stronger during disease progression.

The proposed model still has two limitations, as follows: One is that this work ignores the causal dynamic connections between paired brain regions. The dynamic characteristics are indicative of cognitive and emotional brain activities, which play an important role in detecting the early stage of AD and understanding the pathological mechanisms. In the next work, we will explore the time delay properties of fMRI among brain regions to bridge the dynamic causal connections with biological interpretability. Another is that the input data only concentrates on the fMRI while ignoring other complementary information. Since diffusion tensor imaging (DTI) can characterize the microstructural information, it can enhance the BEC's construction performance and make biological analyses. In the future, we will utilize the GCN and combine it with fMRI to extract complementary information for effective connectivity construction.

\section{Conclusion}
\label{s6}
In this paper, we propose a brain imaging-to-graph generation (BIGG) framework to estimate the brain's effective connectivity from 4-dimensional fMRI in an end-to-end manner.
The proposed framework is based on diffusion models to gradually remove noise from a Gaussian sample with several successive denoising steps. Each denoising step is modeled with a transformer-based GAN to translate the noise and conditional fMRI to the clean sample and effective connectivity. The hierarchical structures in the generator and the discriminator enhance noise removal, making the generation more high-quality, diverse, and efficient.
Results from the ADNI datasets prove the feasibility and efficiency of the proposed model. The proposed model not only achieves superior prediction performance compared with other shallow and deep-learning methods but also identifies MCI-related causal connections for better understanding pathological deterioration and discovering potential MCI biomarkers.

\section*{Acknowledgement}
This work is supported by the Natural Science Foundation of Hubei Province (2023AFB004).
%This work was supported by the National Natural Science Foundations of China under Grant 62172403, the Distinguished Young Scholars Fund of Guangdong under Grant 2021B1515020019, the Excellent Young Scholars of Shenzhen under Grant RCYX20200714114641211, and Shenzhen Key Basic Research Projects under Grant JCYJ20200109115641762.

\end{document}